\documentclass[acmtog]{acmart}
\acmSubmissionID{1992}

\usepackage[ruled]{algorithm2e} 
\usepackage{multirow}
\usepackage{multicol}
\usepackage{enumitem}

\AtBeginDocument{%
  }

\copyrightyear{2025}
\acmYear{2025}
\setcopyright{cc}
\setcctype{by}
\acmConference[SA Conference Papers '25]{SIGGRAPH Asia 2025 Conference Papers}{December 15--18, 2025}{Hong Kong, Hong Kong}
\acmBooktitle{SIGGRAPH Asia 2025 Conference Papers (SA Conference Papers '25), December 15--18, 2025, Hong Kong, Hong Kong}\acmDOI{10.1145/3757377.3763959}
\acmISBN{979-8-4007-2137-3/2025/12}

\begin{document}

\title{SPGen: Spherical Projection as Consistent and Flexible Representation for Single Image 3D Shape Generation}

\author{Jingdong Zhang}
\orcid{0009-0008-6668-1140}
\affiliation{%
  \institution{Texas A\&M University}
  \city{College Station}
  \state{Texas}
  \country{USA}
}
\email{jdzhang@tamu.edu}

\author{Weikai Chen}
\authornote{Corresponding author.}
\orcid{0000-0002-3212-1072}
\affiliation{%
  \institution{LightSpeed Studios}
  \country{USA}
}
\email{weikaichen@global.tencent.com}

\author{Yuan Liu}
\orcid{0000-0003-2933-5667}
\affiliation{%
  \institution{Hong Kong University of Science and Technology}
  \city{Hong Kong}
  \country{China}
}
\email{yuanly@ust.hk}

\author{Jionghao Wang}
\orcid{0009-0002-9683-8547}
\affiliation{%
  \institution{Texas A\&M University}
  \city{College Station}
  \state{Texas}
  \country{USA}
}
\email{jionghao@tamu.edu}

\author{Zhengming Yu}
\orcid{0009-0003-0553-8125}
\affiliation{%
  \institution{Texas A\&M University}
  \city{College Station}
  \state{Texas}
  \country{USA}
}
\email{yuzhengming@tamu.edu}

\author{Zhuowen Shen}
\orcid{0009-0009-3023-5816}
\affiliation{%
  \institution{Texas A\&M University}
  \city{College Station}
  \state{Texas}
  \country{USA}
}
\email{mickshen@tamu.edu}

\author{Bo Yang}
\orcid{0009-0008-1798-4268}
\affiliation{%
  \institution{Waymo}
  \country{USA}
}
\email{yangbo@waymo.com}

\author{Wenping Wang}
\orcid{0000-0002-2284-3952}
\affiliation{%
  \institution{Texas A\&M University}
  \city{College Station}
  \state{Texas}
  \country{USA}
}
\email{wenping@tamu.edu}

\author{Xin Li}
\orcid{0000-0002-0144-9489}
\affiliation{%
  \institution{Texas A\&M University}
  \city{College Station}
  \state{Texas}
  \country{USA}
}
\email{xinli@tamu.edu}

\renewcommand{\shortauthors}{Zhang et al.}

\begin{abstract}
Existing single-view 3D generative models typically adopt multiview diffusion priors to reconstruct object surfaces, yet they remain prone to inter-view inconsistencies and are unable to faithfully represent complex internal structure or nontrivial topologies.
In particular, we encode geometry information by projecting it onto a bounding sphere and unwrapping it into a compact and structural multi-layer 2D Spherical Projection (SP) representation. Operating solely in the image domain, SPGen offers three key advantages simultaneously: (1) \emph{Consistency.} The injective SP mapping encodes surface geometry with a single viewpoint which naturally eliminates view inconsistency and ambiguity; (2) \emph{Flexibility.} Multi-layer SP maps represent nested internal structures and support direct lifting to watertight or open 3D surfaces;
(3) \emph{Efficiency.} The image-domain formulation allows the direct inheritance of powerful 2D diffusion priors and enables efficient finetuning with limited computational resources.
Extensive experiments demonstrate that SPGen significantly outperforms existing baselines in geometric quality and computational efficiency. 

\end{abstract}

\begin{teaserfigure}
  \includegraphics[width=\textwidth]{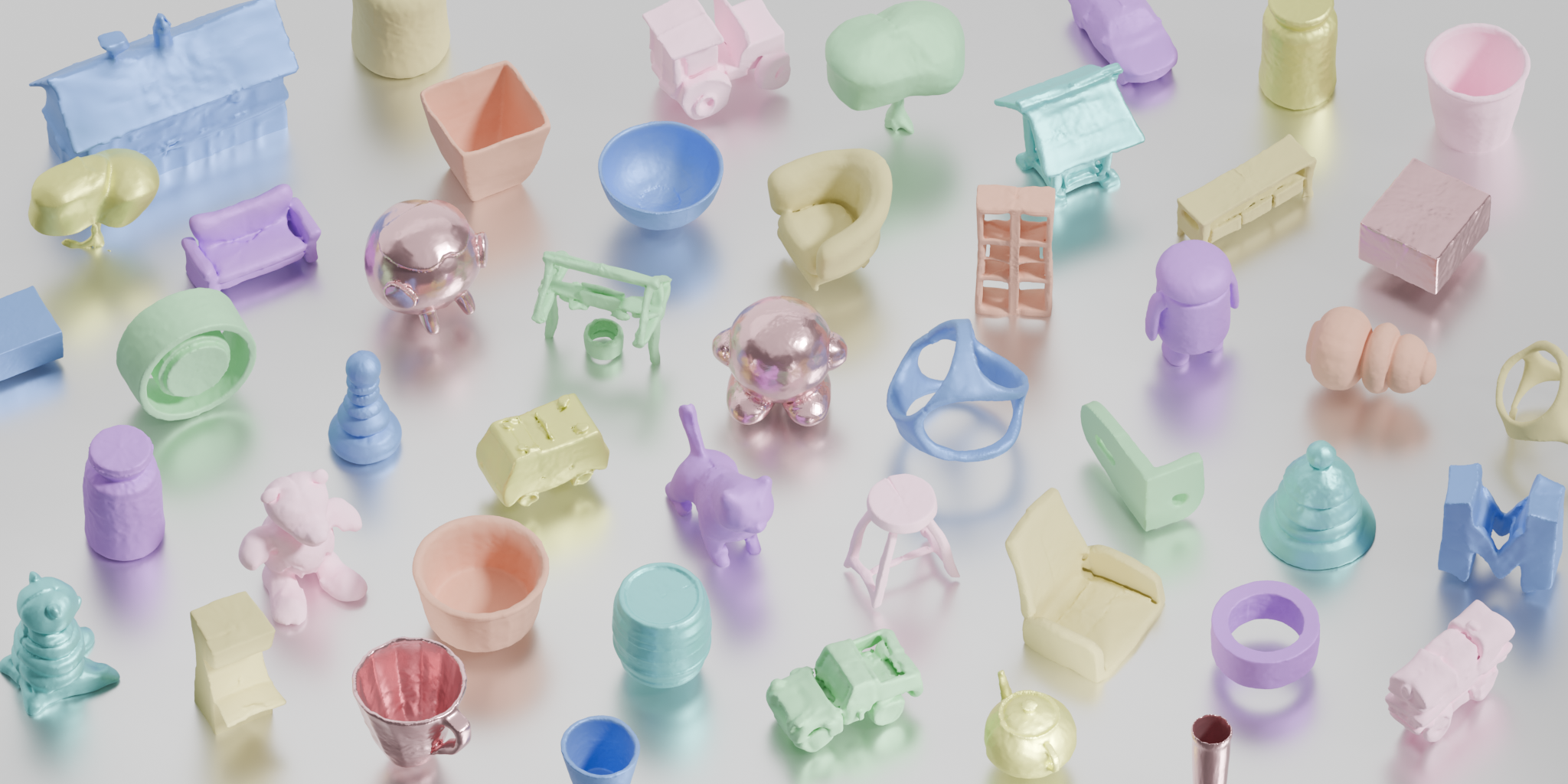}
  \caption{We present SPGen, a powerful generative model creating consistent 3D shapes with flexible topology from single view images in seconds.}
  \label{fig:teaser}
\end{teaserfigure}

\received{20 February 2007}
\received[revised]{12 March 2009}
\received[accepted]{5 June 2009}

\maketitle

\section{Introduction}


\par High-quality 3D asset generation is critical for applications spanning from AR/VR, robotics, industrial design to digital content creation. Conventional modeling workflows are often manual, time-consuming, and require specialized expertise. 
Recent progress in deep generative models~\cite{kingma2013auto,goodfellow2014generative,van2017neural,ho2020denoising,rombach2022high} has substantially advanced the automation and accessibility of this process. By harnessing large-scale datasets and strong priors from pretrained networks, these models\cite{liu2023zero,lin2023magic3d,hong2023lrm,zhang2024clay,long2024wonder3d,chen2024meshanything} enable the synthesis of high-fidelity, semantically meaningful 3D geometry from limited inputs, such as a single image or text, thus facilitating scalable and efficient 3D content creation.


\par Existing 3D generative methods can be broadly categorized based on their intermediate representations. Geometry-based methods~\cite{chen2024meshxl,hao2024meshtron,chen2024meshanything,siddiqui2024meshgpt,alliegro2023polydiff,yu2024surf,zhang2024clay,hong2023lrm,tochilkin2024triposr,li2023instant3d,wu2024direct3d} directly synthesize 3D structures such as point clouds, signed distance fields (SDFs), or explicit mesh faces by using diffusion models, causal transformers, or large reconstruction networks. While effective, unlike the abundance of image or language data, these methods are constrained by the scarcity and noisiness of 3D data, which hinders scalability and requires heavy data preprocessing.

\par To mitigate these challenges, image-based approaches~\cite{richter2018matryoshka,liu2023zero,long2024wonder3d,liu2023syncdreamer,xu2024instantmesh,zhang2018learning,elizarov2024geometry,yan2024object} generate 3D content via intermediate multiview images, geometry image, uv atlas or spherical projection, from which geometry is recovered using differentiable rendering, feed-forward networks, etc. Though these approaches can leverage powerful pretrained 2D priors, they are suffering from several limitations respectively, multiview images are usually lack of strict view consistency and geometric coherence, geometry images and uv atlas are limited by non-unique cuttings and mappings, which burden the model with extensive boundary stitching and hamper scalable training, while simple spherical projection is suffering from severe self-occlusions. These issues either degrade the qualities of restored geometry or limit scalable training on large-scale datasets.



\par To address the limitations of existing 3D generative methods, we propose SPGen, a novel scalable framework that generates high-quality meshes based on \emph{multi-layer Spherical Projection} (SP) maps.
Concretely, given a normalized 3D object, we enclose it within a unit sphere and cast rays from the origin outward through each point on the spherical surface. For each ray, the intersection information, such as depth, is recorded at the corresponding point on the sphere. The sphere is subsequently projected onto a 2D image plane, forming what we term a {Spherical Projection (SP)} map. For general objects exhibiting self-occlusion or nested internal geometry, we trace multiple intersections along each ray and store them sequentially in multi-layer SP maps, thereby capturing fine-grained spatial structure beyond the external surface.
This SP-based formulation underpins our image-centric generation pipeline and confers three key advantages simultaneously.
(1) \emph{Consistency.} The SP map is a naturally view-consistent representation encoding 360-degree geometry. The mapping of valid pixel to surface point is an injective function, which eliminates potential view conflicts.
(2) \emph{Flexibility.} Multi-layer SP maps enables faithful representation of geometries with varying resolution and topology. Notably, both watertight and open surfaces, as well as layered internal structures, can be directly reconstructed from multi-layer SP maps.
(3) \emph{Efficiency.} As SP maps are structured 2D representations, we can finetune powerful pretrained diffusion backbones such as SDXL~\cite{podell2023sdxl} with limited resource consumptions, and meanwhile inherit strong prior knowledge including locality, semantic, implicit symmetry and repetition patterns instead of learning from scratch.

\par Moreover, we introduce tailored components to address the challenges when applying general generative pipelines specifically on SP maps. We observe that errors of SP maps during training mainly gather at geometric boundaries, thus we propose to adopt a composition of geometry regularization at the boundaries to enhance the SP map qualities. For multi-layer SP generation, we leverage layer-wise self-attention to enforce alignment between interior and exterior layers. After denoising, we perform unprojections from SP maps to 3D spaces to obtain a dense 3D point cloud, followed by mesh extraction via a lightweight feed-forward network. SPGen can generate 3D meshes with high geometric quality in seconds. Experimental results demonstrate that SPGen achieves superior performance compared to prior methods, despite requiring significantly less training overhead.

\par In summary, our contributions are three-fold:
\begin{itemize}[nosep]
\setlength{\itemsep}{0pt}
\setlength{\parskip}{0pt}
    \item We propose to use multi-layer Spherical Projection (SP) maps as compact and structural representations for 3D shape generation. The SP maps encode the whole surface geometry through injective mappings, enabling view-consistent and topology-flexible geometry reconstruction.
    \item We present a novel generation pipeline SPGen. By efficiently finetuning the powerful image diffusion model with limited resources to inherit rich prior knowledge, and incorporating specifically designed geometry regularization and layer-wise self-attention, SPGen could generate high-quality 3D meshes in seconds via single image conditioning.
    \item Our proposed method achieves state-of-the-art performance on public benchmarks with significantly lower training overhead, indicating the effectiveness and efficiency of SPGen.
\end{itemize}

\begin{figure*}[t]
  \centering
   \includegraphics[width=1.0\linewidth]{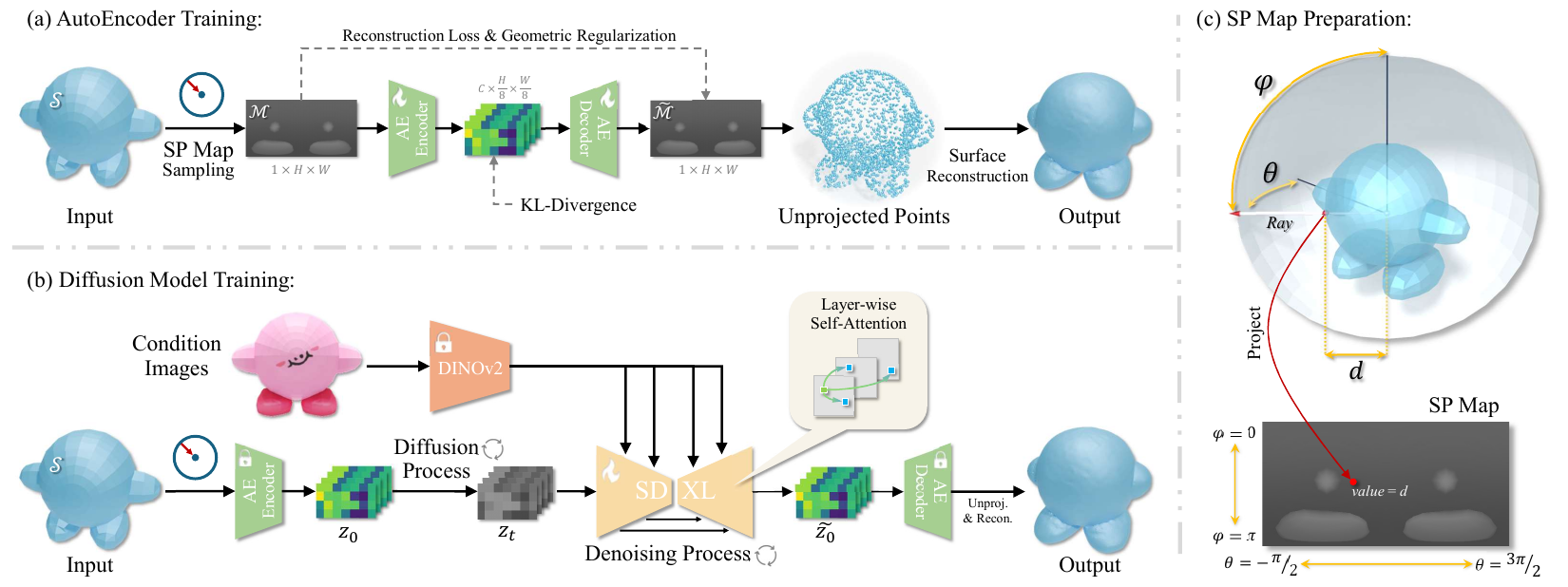}
   \vspace{-6mm}
   \caption{\textbf{Illustration of SPGen.} (a) The AutoEncoder (AE) Training Pipeline: we train the AutoEncoder on projected Spherical Projection (SP) maps. We introduce geometry regularization to aid the reconstruction process, and KL divergence is applied to regularize the latent distribution. After the SP map is reconstructed, we unproject points into 3D space and extract the surface. (b) The Diffusion Model Training Pipeline: we use the finetuned AutoEncoder to produce latent and train the denoise UNet with conditioning image embeddings from DINOv2, except from the standard self-attention guiding the diffusion process, we also introduce layer-wise self-attention among multi-layer SP maps. The denoised latent is fed into AE decoder and produces the final mesh output similarly. (c) The detailed preparation of one layer of SP Map: the object mesh is normalized and placed at the center of a union sphere. We cast rays from the origin and record the depth value $d$ on the SP map parameterized by azimuth angle $\theta$ and polar angle $\varphi$. } 
   \vspace{-3mm}
   \label{fig:main}
\end{figure*}

\section{Related Works}

\subsection{3D Shape Creation with Geometric Representations}
\par Creating high-quality 3D shapes from single view input has received a lot of attention recently~\cite{liu2023zero,lin2023magic3d,poole2022dreamfusion,tang2023dreamgaussian,li2023instant3d,hong2023lrm,tochilkin2024triposr,xu2024instantmesh,long2024wonder3d,tang2025lgm,wang2025crm,hu2024x,yu2024surf,wang2025disentangled,chen2024meshxl,hao2024meshtron,chen2024meshanything,siddiqui2024meshgpt,alliegro2023polydiff,zhang2024clay,wu2024direct,wu2018learning,richter2018matryoshka,zhang2018learning,wang2025pdt}. Based on the representation adopted, these works can be roughly divided into two groups, geometry-based and image-based. The former usually takes point clouds, Signed Distance Fields (SDF) fields, or directly uses mesh faces as surface representations. For point clouds, Point-E~\cite{nichol2022point} tries to denoise point cloud directly by a transformer-based diffusion model, CLAY~\cite{zhang2024clay}, Direct3D~\cite{wu2024direct3d} further enhanced this idea by applying an autoencoder to compress the point clouds and acquire compact and representative latents, which makes it easier to scale-up the model. Taking SDF as a shape representation, SDF-StyleGAN~\cite{zheng2022sdf} and SDF-3DGAN~\cite{jiang2023sdf} extend Generative Adversarial Network (GAN) from 2D images to 3D shapes. LAS-Diffusion~\cite{zheng2023locally} introduces a two-stage pipeline generating refined SDFs from coarse voxel occupancy fields. SurfD~\cite{yu2024surf} proposed to generate Unsigned Distance Fields (UDF) to represent open surfaces. Different from denoising SDF~\cite{li2023diffusion,zheng2023locally,yariv2024mosaic}, Large Reconstruction Models (LRM)~\cite{hong2023lrm,tochilkin2024triposr,li2023instant3d} directly learn SDF from single or multiple input images by feed-forward transformers. Another bunch of works directly take mesh faces as representations, PolyDiff~\cite{alliegro2023polydiff} trains a diffusion model to denoise mesh faces while~\cite{siddiqui2024meshgpt,chen2024meshxl,chen2024meshanything} encodes each face as a token and predicts the next token autoregressively to predict the whole mesh. Despite the achievements of geometry-based methods, these methods have higher requirements for the quality and form of data, for example, autoregressive methods constrain the number of mesh faces and SDF methods need to calculate SDFs for water-tight objects only, which requires more complex data preprocessing processes and limits the scaling-up of these methods.

\subsection{3D Shape Creation from Image-based Representations}

\par To solve the shortcomings of geometry-based representations, and utilize the strong priors stored in powerful 2D generative models trained with billions of data, another bunch of works takes images as the representation to record the geometry. Matryoshka Network~\cite{richter2018matryoshka} proposes to mark the space between each “entry–exit” depth pair in its six axis-aligned stacks as occupied, fuses the results into a voxel volume to restore both external and internal of an object completely, while Genre~\cite{zhang2018learning} represents the outer surface by a single-layer spherical depth map, and takes the reconstruction from single image as a spherical map inpainting task. However, their restored geometry qualities are limited by the resolution bottleneck of voxel grids during surface extraction. Different from them, another bunch of works take advantage of geometry images~\cite{gu2002geometry,elizarov2024geometry} or uv atlas~\cite{yan2024object} to unfold the surface geometries to image charts, these representations require non-unique cuttings and mappings, which burden the model with extensive boundary stitching and hamper scalable dataset preparation.

\par Recently, multiview-based representations are also quickly thriving, DreamFusion~\cite{poole2022dreamfusion} proposes Score Distillation Sampling (SDS) to distill from image diffusion models and extract surface geometry by differentiable rendering~\cite{mildenhall2021nerf,laine2020modular,kerbl20233d,wang2021neus,huang20242d,guedon2024sugar,yariv2021volume,shen2024solidgs} without training on 3D datasets, and ~\cite{lin2023magic3d,tang2023dreamgaussian} further improves the results. However, these methods are computationally expensive and usually suffer from blurred details. Some works focus on generating multi-view consistent images from one input image and therefore reconstruct the geometry~\cite{liu2023zero,long2024wonder3d,liu2023syncdreamer,xu2024instantmesh,shi2023mvdream,liu2024one}, among them, Zero123~\cite{liu2023zero} firstly tries to incorporate camera as conditions and SyncDreamer~\cite{liu2023syncdreamer} introduces spatial attention to align views and extract surfaces by~\cite{wang2021neus}. Wonder3D~\cite{long2024wonder3d} enhances the performance by introducing cross-domain diffusion for both RGB images and normal maps, and Zero-1-to-G~\cite{meng2025zero} extends it to incorporate multiple Gaussian attributes. ~\cite{xu2024instantmesh,wang2025crm,tang2025lgm,xu2024grm,wu2024direct} are combinations of both multiview generation process and feed-forward reconstruction process, while CRM~\cite{wang2025crm} focuses on the strong connections between canonical views and triplane features, ~\cite{tang2025lgm,xu2024grm} focus on feed-forward Gaussians reconstruction, and~\cite{wu2024direct} focuses on multi-view depth generation. However, these methods mentioned above heavily rely on the quality of multi-view image generation, and strict constraints of view consistency are not guaranteed by solely applying cross-attention as a soft constraint, consequently degrading the geometry quality when inconsistencies occur. Besides, multi-view images fail to represent objects with severe self-occlusions or internal layers. Differently, we propose to use multi-layer Spherical Projection to record the whole object geometry and serve as a consistent and coherent representation.

\section{Methodology}
\par The detailed design of our proposed SPGen pipeline is shown in Fig~\ref{fig:main}. Firstly, we extract the SP maps from object meshes. After the SP maps are prepared, we finetune the image AutoEncoder on SP maps to obtain compact latents. Then we finetune the latent diffusion model to generate multi-layer SP maps. Finally, the SP maps are used to reconstruct high-quality shapes by Poisson reconstruction or UDF reconstruction to represent water-tight or open surfaces.

\subsection{Spherical Projection}
\label{sec:sp}
\par Current methods taking advantage of pretrained image generative models usually adopt multiview images as representations, since multiview images comprehensively cover the exterior of the object and project surface points to 2D domain. However, these projections are not simple injective functions due to the overlapping of adjacent views, which consequently leads to ambiguity during the generation process. Since the object surface is usually a complex 2D manifold, we try to find an injective projection that maps it to a structural 2D domain that could be handled by image generative models. Therefore, we propose Spherical Projection (SP) as the shape representation.

\par Adopting SP maps as a panorama for scene-level generation is a common solution~\cite{wang2023360,feng2023diffusion360,lu2024autoregressive,yan2024horizon,hara2022spherical}. Differently, we are introducing SP maps as the geometry representations for shape generation. As shown in Fig.~\ref{fig:main} (c), we firstly cast a ray from the origin along the radial direction with azimuth angle $\theta$ and polar angle $\varphi$. When the ray intersects with a surface point $\mathcal{P} \in \mathbb{R}^{3}$, we calculate the distance that the ray travels as $d = \| \mathcal{P} \|_2$. In this simple way, we can use equirectangular projection~\cite{hara2022spherical} $\operatorname{F}{(\mathcal{P})}: \mathbb{R}^{3} \to \mathbb{R}^{2}$, to map a 3D point to the 2D domain parameterized by $\theta$ and $\varphi$. By recording the corresponding $d$ at each point $\left(\theta, \varphi \right)$, the geometry is recorded on the SP map. To acquire the original point position, we simply perform the conversion from spherical coordinates to Euclidean coordinates:
\vspace{-1mm}
\begin{equation}
\mathcal{P}
= \operatorname{F^{-1}}(\theta, \phi) = 
\begin{bmatrix}
\sin\phi \cos\theta \\
\sin\phi \sin\theta \\
\cos\phi
\end{bmatrix}
d .
\label{eq:map}
\end{equation}

\par For rays that intersect with the mesh surface more than once, we record all the intersection positions, and project them onto SP maps reversely, i.e., starting from the outermost intersection, until we reach the maximum recording depth or there are no more intersections. In this manner, we not only solve the self-occlusion issue, but also empower the SP maps to represent the inner structures of a complex object. As shown in Algo.~\ref{alg:SPGen}, assume that we want to prepare $k$ layers of SP maps $\{\mathcal{M}^1, \mathcal{M}^2, \dots, \mathcal{M}^k\}$ for mesh surface $\mathcal{S}$, for each ray $\mathcal{R}_{i}$, we need to perform ray-mesh intersection penetrating all layers and record the intersection points $\{ \mathcal{P}_i^0, \mathcal{P}_i^1, \dots, \mathcal{P}_i^k \}$. If there is no more intersections after layer $j^{\prime}$, where $j^{\prime} \le k$, then $ \mathcal{P}_i^{j^{\prime}}, \dots, \mathcal{P}_i^k$ are set as \textit{NULL}. Then we loop through each layer of SP map $\mathcal{M}^j$ reversely and record the depth values $d = \|P_i^{k-j}\|_2$ of valid points on the map.

\begin{small} 
\begin{algorithm}[t]
\SetAlgoNoLine
\KwIn{The object surface $\mathcal{S}$}
\KwOut{$k$-layer SP maps $\mathfrak{M} = \{\mathcal{M}^1, \mathcal{M}^2, \dots, \mathcal{M}^k\}$ storing depth}

Initialize SP map layers as empty\;
Initialize rays $\mathfrak{R} = \{\mathcal{R}_1, \mathcal{R}_2, \dots, \mathcal{R}_n\}$, uniformly sample $(\theta, \phi)$\;
\For{each point $\mathcal{R}_i \in \mathfrak{R}$}{
    $\{ \mathcal{P}_i^0, \mathcal{P}_i^1, \dots, \mathcal{P}_i^k \}$ = \textbf{RayMeshIntersection}$(\mathcal{R}_i, \mathcal{S}, k)$\;
    \textit{step} = 0\;
    \For{j in range$(k)$}{
        \If{$P_i^{k-j}$ is not NULL}{
            $\mathcal{M}^{step} = d = \|P_i^{k-j}\|_2$\;
            \textit{step} += 1\;
            }
        }
}
\caption{Multi-layer SP Map Preparation}
\label{alg:SPGen}
\end{algorithm}
\end{small}

\par After the SP maps are prepared, we finetune the AutoEncoder and Diffusion Model to generate 3D shapes based on them, which will be explained in detail in the following sections.

\subsection{Generation Pipeline}
\label{sec:gen}

\subsubsection{Preliminaries}
\label{sec:pre}
\par First, we are going to introduce our training pipeline. Our generation pipeline is built upon SDXL~\cite{podell2023sdxl}, leveraging the strong prior from pretraining. Specifically, this stable diffusion pipeline is composed of a set of Kullback–Leibler divergence regularized (KL-regularized) Autoencoder $\Psi_{\mathcal{E}}, \Psi_{\mathcal{D}}$ and a large-scale denoising UNet $\Theta$. The AutoEncoder compresses high-resolution input $\mathcal{M}$ to compact latent $z_0$, and this process is optimized by jointly minimizing reconstruction error and regularizing latent distribution:
\begin{equation}
\begin{split}
z_0 &\sim \mathcal{Q}\left({z} \mid \mathcal{M}\right), \\
L_{recon} &= \mathbb{E}_{\mathcal{M}}\Big[\left\|\mathcal{M} - \Psi_{\mathcal{D}}\left(\Psi_{\mathcal{E}}\left(\mathcal{M}\right)\right)\right\|\Big] \\
&\quad + \lambda \cdot \mathbb{E}_{\mathcal{M}}\Big[D_{\mathrm{KL}}\left(\mathcal{Q}\left({z} \mid \mathcal{M}\right) \| \mathcal{N}(0, I)\right)\Big] ,
\end{split}
\label{eq:aeloss}
\end{equation}
\noindent where $\mathcal{Q}$ is the output distribution of $\Psi_{\mathcal{E}}$ and $\lambda$ is the control coefficient of regularization strength. After $\Psi_{\mathcal{E}}, \Psi_{\mathcal{D}}$ are trained, $z_0$ is generated accordingly and a noise scheduler gradually adds Gaussian noise to it over $T$ time steps: 
\begin{equation}
z_{t}=\sqrt{\alpha_{t}} z_{0}+\sqrt{1-\alpha_{t}} \epsilon, \quad \epsilon \sim \mathcal{N}(0, I) ,
\label{eq:add_noise}
\end{equation}
\noindent where ${z_t}$ is the noisy version at time step $t < T$, and $\mathbf{\alpha_t}$ is the noise schedule coefficient. Then the denoise UNet is trained to parameterize the reverse diffusion process by predicting the noise $\epsilon_\Theta({z_t}, t)$ of time step $t$. Finally, the optimization process is achieved by minimizing the mean squared error (MSE) between the predicted noise and the actual noise:
\begin{equation}
L_{diff} =\mathbb{E}_{z_{0}, \epsilon, t}\Big[\left\|\epsilon-\epsilon_{\Theta}\left(z_{t}, t\right)\right\|^{2}\Big] .
\label{eq:diffloss}
\end{equation}

After training, synthesis data is generated by reversing the noise-adding process in latent space, and the denoised latent is fed to $\Psi_{\mathcal{D}}$ to produce final results.

However, since the diffusion model is only trained on RGB data for image synthesis, which mainly focuses on better perceptual quality and lacks the generalization ability on Sp depth maps, solely inference on pretrained pipeline results in poor geometry quality. Thus, we propose to finetune the whole pipeline and we will illustrate the details in the following sections.

\subsubsection{Layer-wise Self Attention}
\label{sec:lsa}
\par Applying attention to associate multiple predicting objectives is a popular technique in recent works~\cite{shi2023mvdream,long2024wonder3d,fu2025geowizard,ye2022inverted,zhang2025bridgenet}. In our pipeline, we need to constrain the generated layers to have reasonable relative positions in space, and self-intersections or floating artifacts are therefore avoided.

\par Considering the intermediate hidden states $\{m^1, m^2, \dots, m^k\}, m^j \in \mathbb{R}^{C \times h \times w}$ generated by UNet parameters $\Theta$ that corresponding with SP maps $\{\mathcal{M}^1, \mathcal{M}^2, \dots, \mathcal{M}^k\}$. We first flatten them by $\operatorname{Flat}(m^j): \mathbb{R}^{C \times h \times w} \to \mathbb{R}^{C \times (hw)}$ and concatenate them along the spatial dimension:
\vspace{-1mm}
\begin{equation}
\bar{m} = \operatorname{Concat}\left( \left[ \operatorname{Flat}(m^1), \dots, \operatorname{Flat}(m^k) \right], dim=-1 \right) ,
\label{eq:concat}
\end{equation}
\noindent where $\operatorname{Concat}(\cdot)$ represents the concatenation operation. Followingly, we perform the standard self-attention on $\bar{m}$:
\begin{equation}
\text { Attention }(Q, K, V)=\operatorname{softmax}\left(\frac{Q K^{T}}{\sqrt{C_{a}}}\right) \cdot V ,
\label{eq:attn}
\end{equation}
\noindent where $Q$, $K$ and $V$ are produced by linear projections from $\bar{m}$, and $C_{a}$ represents the projected dimension for attention. In this way, layer relations are modeled, leading to a more accurate geometry during the denoising process.

\begin{figure}[t]
  \centering
  \includegraphics[width=0.78\linewidth]{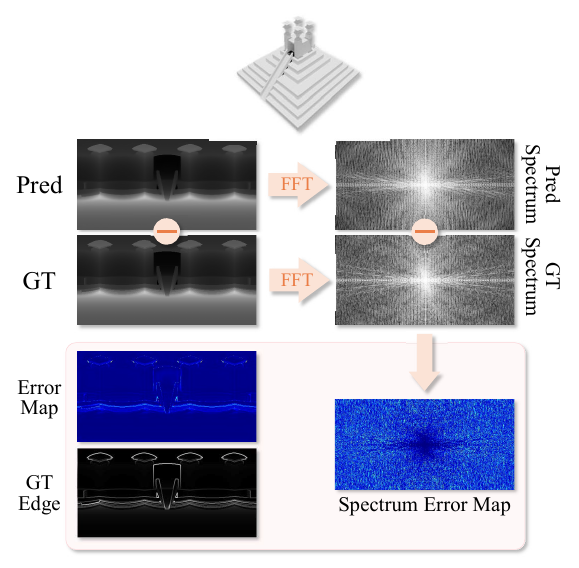}
  \vspace{-3mm}
  \caption{\textbf{Visualization of error distributions.} We visualize the edge map, spectrum map, and corresponding error maps of prediction and groundtruth (GT). The error distribution is highly aligned with the edge distribution, indicating a large amount of error falls on areas with large image gradients. This observation is consistent with the phenomenon in the spectral domain, in which high-frequency component dominates the error distribution.
  } 
  \vspace{-6mm}
  \label{fig:comp_geo}
\end{figure}

\subsubsection{Geometry Regularization}
\label{sec:losses}
\par In Sec.~\ref{sec:pre}, we analyzed the necessity of fine-tuning the diffusion model. However, simply transplanting the methods of training image diffusion is suboptimal since these methods (such as perceptual loss~\cite{johnson2016perceptual}) do not contribute to better geometry quality. Among the pipeline components, finetuning the AutoEncoder $\Psi$ is the most tricky part, since the input and output are SP maps, which have very different distributions from the original RGB image, on the contrary, the KL regularization forces the latent distribution to be closer to the original standard normal distribution, which poses a great challenge to the training process. 

\par During training, we observed that if we only apply L1 distance as the reconstruction loss, the reconstructed results are not satisfying with blurred details and noisy surfaces. We study the output by analyzing error maps. As shown in Fig.~\ref{fig:comp_geo}, we observe that the error in the spatial domain is concentrated on the edge of the SP Map, which is the high-frequency component of an image. This concentration of error occurs because standard reconstruction losses, like L1 distance, average the error over all pixels. Since edges and details are a small fraction of total pixels, their error has minimal impact on the overall loss. The model therefore prioritizes optimizing large, smooth areas, causing the high-frequency details crucial for accurate geometry to become blurred or misplaced. Thus, we further visualize the spectrum produced by the Fast Fourier Transform (FFT), which consistently shows that the error mainly exists in the four corners of the spectrum, which is where the high-frequency components exist, while the center where the low-frequency components are located is darker, indicating that the error is smaller. This is consistent with the conclusion in~\cite{jiang2021focal}.

\par Inspired by this observation, we propose two regularization losses to enhance the geometry quality. First, we directly strengthen supervision of image boundaries, since the pixel L1 loss will tend to punish overall deviation, while the pixel ratio of the edge is too small to be fully supervised, we extract a hard boundary mask $\mathcal{B}$ with margin by Sobel operator and dilatation operator: $\mathcal{B} = \operatorname{Dilate}\left(\operatorname{Sobel}(\mathcal{M}) \right)$, then we use $\mathcal{B}$ to mask out the boundary pixels and impose greater punishment on them individually:
\begin{equation}
\begin{split}
L_{edge} & = \mathbb{E}_{\mathcal{M}}\Big[ \mu \mathcal{B} \cdot  \left\|\mathcal{M} - \Psi\left(\mathcal{M}\right)\right\| \\
& + (1 - \mu) (1 - \mathcal{B}) \cdot \left\|\mathcal{M} - \Psi\left(\mathcal{M}\right)\right\| \Big] ,
\end{split}
\label{eq:edge_loss}
\end{equation}
\noindent where $\Psi = (\Psi_{\mathcal{D}} \circ \Psi_{\mathcal{E}})$, and $\mu$ is a control coefficient of the strength, by applying a larger weight $\mu$ to the loss calculated within this masked region $\mathcal{B}$, we compel the AutoEncoder to pay closer attention to these critical areas. This targeted penalty prevents the model from smearing details across edges and results in a more precise reconstruction of sharp geometric contours and object silhouettes.

Second, we enhance the high-frequency component from the spectral domain. We first perform Fast Fourier Transform (FFT) to both prediction and groundtruth, denoting as $\mathcal{M}_s = \operatorname{FFT}(\mathcal{M})), \tilde{\mathcal{M}_s} = \operatorname{FFT}(\operatorname{\Psi}(\mathcal{M})))$, decomposing the SP map into its constituent frequencies. And then impose a high pass filter $\mathcal{H}$ on the spectrum, which is a circular mask that covers the central area where low-frequency components are located, and only allows high-frequency components to compute loss, we separately calculate the L1 distance on principal value of argument and modulus respectively:
\begin{equation}
\begin{split}
L_{spec} & = \mathbb{E}_{\mathcal{M}} \Big[ \mathcal{H} \cdot \left\| \operatorname{Arg}( \mathcal{M}_s ) - \operatorname{Arg}( \tilde{\mathcal{M}_s} )\right\| \\
& + \zeta \mathcal{H} \cdot \left\| \| \mathcal{M}_s \|_2 - \| \tilde{\mathcal{M}_s}\|_2 \right\| \Big] ,
\end{split}
\label{eq:spec_loss}
\end{equation}
\noindent where $\zeta$ is also a coefficient. By separately penalizing discrepancies in both the phase and magnitude of these frequencies, the model significantly reduces surface noise and improves the crispness of the final reconstructed geometry. With $L_{recon}$ from Eq.~\ref{eq:aeloss}, the total loss for training the AutoEncoder can be written as:
\begin{equation}
L = L_{recon} + \alpha L_{edge} + \beta L_{spec} .
\label{eq:edge}
\end{equation}

\subsubsection{Surface Extraction}
\label{sec:remesh}
\par After the SP maps are generated, we unproject points to 3D space by Eq.~\ref{eq:map} and extract the surface from the point cloud. For watertight objects, we simply perform Poisson reconstruction. We sample oriented point clouds from groundtruth meshes and use them to train a light-weight 3D-Unet as a normal estimator, which predicts a unit normal vector for each point. On this basis, the gradient field is calculated and the surface is reconstructed. For open surfaces, we follow~\cite{yu2024surf}, use their pretrained point-cloud-to-UDF AutoEncoder to predict the UDF in space, and the implicit surface is extracted by MeshUDF~\cite{guillard2022meshudf}. The reconstruction process is exceptionally fast while remaining cost-efficient, with high-quality geometry generated from SPGen, we can obtain accurate and detailed meshes suitable for downstream tasks such as editing, rendering, simulation, etc.

\vspace{-1mm}
\section{Experiments}

\subsection{Experimental Setting}
\subsubsection{Dataset}
We refer to the criteria in~\cite{chen2024meshanything,long2024wonder3d} to filter the Objaverse dataset~\cite{deitke2023objaverse} by removing low-quality or scene-level meshes and acquire around 160k objects as our whole training split. Before training, we follow~\cite{long2024wonder3d} to render the multi-view image for reference. We also picked 1993 objects out of the training indices as our validation split on Objaverse. We normalize the scales of object meshes to the range of $\left[-0.5, 0.5\right]$, and translate objects to the origin, where the sphere-center is also fixed for scanning SP maps. We then render $4$ layers of SP maps per object with the resolution $256 \times 512$ since we empirically discover 4 layers of SP maps are adequate to cover almost all surface points. For evaluation, we follow prior works to use the Google Scanned Objects (GSO) dataset~\cite{downs2022google} and we randomly choose $30$ shapes consisting of common objects used in daily life. We use the same protocol to render one image as the evaluation input. We further exploit the Deepfashion3D dataset~\cite{zhu2020deep} to illustrate our model capacity on open-surface objects, we follow the setting in~\cite{yu2024surf} to divide the train and test splits. For data from Deepfashion3D, we render $3$ layers of SP maps per object with the resolution $256 \times 512$, and we use the same protocol as~\cite{yu2024surf} to generate sketch for each input view.

\subsubsection{Metrics}
To evaluate the geometry quality of our method, we report Chamfer Distance (CD), Volume IoU and F-Score (with a threshold of 0.1) between the reconstructed mesh and groundtruth mesh. Since the generated meshes are usually placed at different angles, we follow~\cite{huang2025spar3d} and perform brute-force search in rotations to align each predicted mesh with the groundtruth mesh before centering and scaling all meshes to $[-1, 1]$.

\begin{figure*}[t]
  \centering
   \includegraphics[width=1.0\linewidth]{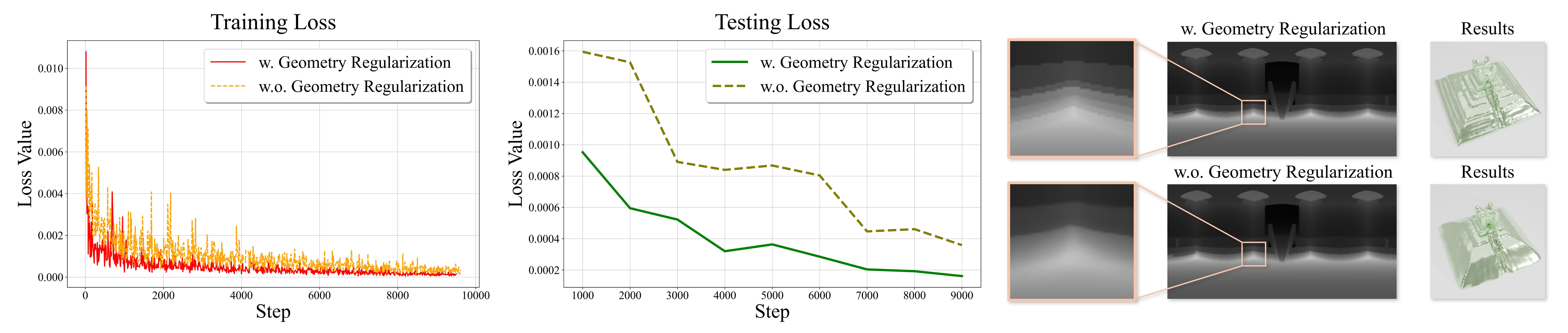}
   \vspace{-7mm}
   \caption{\textbf{Analysis on geometry regularization.} On the left, we compare the loss curve on training and testing splits, and on the right we visualize the geometry details on SP maps and reconstructed mesh surfaces.}
   \vspace{-3mm}
   \label{fig:abs_gr}
\end{figure*}

\subsubsection{Baselines}
We compare to state-of-the-art 3D creation models with single view image (or sketch) as condition, including geometry-based representations Point-E~\cite{nichol2022point}, Shape-E~\cite{jun2023shap}, OpenLRM~\cite{he2023openlrm} (an open-sourced implementation of LRM~\cite{hong2023lrm}) and SurfD~\cite{yu2024surf}. Multi-view based methods Wonder3D~\cite{long2024wonder3d}, CRM~\cite{wang2025crm}, LGM~\cite{tang2025lgm} and InstantMesh~\cite{xu2024instantmesh}. We use their official code implementations and pretrained weights.

\begin{table}[t]
\begin{center}
\setlength{\tabcolsep}{4.9mm}{\scalebox{0.75}{
\begin{tabular}{lcccc}
 \toprule
{Method}& Latency & {CD.}↓ &{Vol. IoU}↑ & {F-Sco. (\%)↑}\\
\midrule
Point-E & $\sim$25s &0.0690 & 0.1953  & 52.23 \\
Shape-E & $\sim$20s &0.0418 & 0.2785  & 64.83 \\
Wonder3D & $\sim$10min  &0.0398 & 0.2930  & 68.82 \\
CRM   & $\sim$18s &0.0264   &0.3374   &74.43 \\
OpenLRM     & $\sim$15s &0.0344   &0.3770   &71.50 \\
LGM     &$\sim$40s  &0.0212   &0.4220   &78.41 \\
InstantMesh &$\sim$35s  &0.0120   &0.4310   &88.84 \\ \midrule
Ours      &6-10s  &\textbf{0.0051}    &\textbf{0.5407}    &\textbf{95.57} \\
\bottomrule
\end{tabular}
}}
\end{center}
 \caption{\textbf{Quantitative comparison on GSO.} Our specific inference time (latency) depends on how many steps we use for denoising and adopting the normal estimator with different sizes.}
 \vspace{-8mm}
\label{tab:1}
\end{table}

\begin{table}[t]
\begin{center}
\setlength{\tabcolsep}{7.3mm}{\scalebox{0.75}{
\begin{tabular}{lccc}
 \toprule
{Method}& {CD.}↓ &{Vol. IoU}↑ & {F-Sco. (\%)↑}\\
\midrule
Wonder3D& 0.0223  & 0.3370  & 73.52 \\
OpenLRM & 0.0237  & 0.3680  & 78.25 \\
LGM & 0.0244  & 0.3110  & 71.60 \\
InstantMesh &0.0314   &0.2890 & 68.72 \\
SurfD   &0.0136   &0.3860 & 82.31 \\ \midrule
Ours (rgb)  &0.0099   &0.4200   &87.16 \\
Ours (sketch) &\textbf{0.0092}    &\textbf{0.4480}    &\textbf{89.35} \\
\bottomrule
\end{tabular}
}}
\end{center}
 \caption{\textbf{Quantitative comparison on DeepFashion3d.}}
 \vspace{-8mm}
\label{tab:2}
\end{table}

\begin{table}[t]
\begin{center}
\setlength{\tabcolsep}{1.3mm}{\scalebox{0.75}{
\begin{tabular}{lcccccccc}
 \toprule
Resolution & \multicolumn{2}{c}{32} & \multicolumn{2}{c}{64} & \multicolumn{2}{c}{128} & \multicolumn{2}{c}{256} \\
\cmidrule(lr){2-3}\cmidrule(lr){4-5}\cmidrule(lr){6-7}\cmidrule(lr){8-9}
 & {CD.}↓ &{Storage}↓ & {CD.}↓ &{Storage}↓ & {CD.}↓ &{Storage}↓ & {CD.}↓ &{Storage}↓ \\
\midrule
 Matryoshka & 7.59 & 10 & 2.43 & 25 & 1.43 &    78  & 0.95  & 261 \\
 UV Mapping & 6.28 & 8   & 2.29  & 32 & 1.16 & 128 & 0.88 & 512 \\
 Ours & 2.66 & 5 & 1.58 & 16 & 0.96 & 56 & 0.85 & 194 \\
\bottomrule
\end{tabular}
}}
\end{center}
 \caption{\textbf{Quantitative comparisons with image-based methods Matryoshka~\cite{richter2018matryoshka} and UV Mapping~\cite{yan2024object}} in terms of representation capacity (Chamfer Distance, CD. $\times 10^{-3}$) and storage efficiency ($KB$) by reconstructing the ground-truth meshes under different image resolutions.}
 \vspace{-6mm}
\label{tab:2.5}
\end{table}

\begin{figure}[t]
  \centering
  \includegraphics[width=1.0\linewidth]{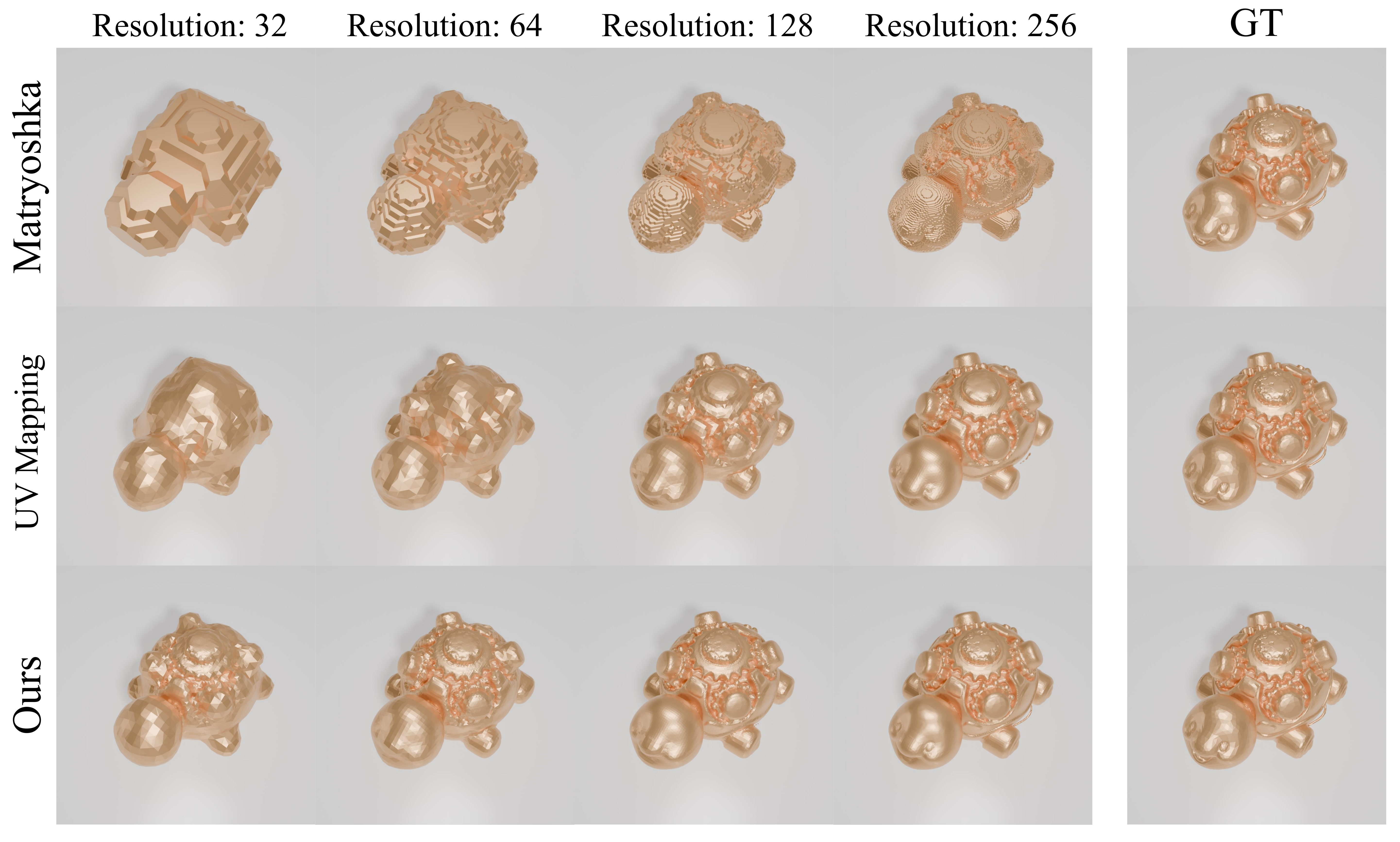}
  \vspace{-6mm}
  \caption{\textbf{Visualization reconstruction quality of Matryoshka~\cite{richter2018matryoshka} and UV Mapping~\cite{yan2024object}.} We reconstruct the ground-truth under different image resolutions, the comparison shows that our SP maps maintain better geometry details and surface qualities.
  } 
  \vspace{-3mm}
  \label{fig:comp_gt_more}
\end{figure}

\begin{table}[t]
\begin{center}
\setlength{\tabcolsep}{6.3mm}{\scalebox{0.75}{
\begin{tabular}{lccc}
 \toprule
{Method}& {CD.}↓ &{Vol. IoU}↑ & {F-Sco. (\%)↑}\\
\midrule
Ours& 0.0051  & 0.5407  & 95.57\\
\textit{w.o.} LSA   &0.0436   &0.2568   &60.75\\
\textit{w.o.} finetuning AE & 0.0610  & 0.2072  & 52.04 \\
\textit{w.o.} finetuning UNet & 0.1742    &0.1034   &27.42 \\
\bottomrule
\end{tabular}
}}
\end{center}
 \caption{\textbf{Ablations on Layer-wise Self-Attention (LSA) and finetuning.}}
 \vspace{-9mm}
\label{tab:3}
\end{table}

\subsubsection{Implementation Details}
We firstly modify the input and output channel of the AutoEncoder from $3$ to $1$ to fit the single-channel depth input, and then finetune with our proposed geometry regularization for 40k iterations with a total batchsize $64$ on all training data from Objaverse. The initial learning rate is set as $1 \times 10^{-4}$ with a cosine annealing learning rate scheduler. After the AutoEncoder is finetuned, we offline generate SP map latents and use them to finetune the denoise UNet. We train the UNet for 80k iterations with a total batchsize $80$ per shape on all data. The initial learning rate is set as $1 \times 10^{-5}$ with warm up for $100 steps$ and annealing learning rate scheduler, and we use DDIM~\cite{song2020denoising} and Euler Ancestral Discrete~\cite{karras2022elucidating} noise scheduler for training and inference respectively. We follow~\cite{wang2023360} and apply circular padding to the SP maps. After the finetuning on Objaverse, we continue to finetune the model on the deepfashion3D training split since the SP maps have different numbers of layers in which exist potential gaps. We train with both sketch and RGB images as conditions to obtain versatile generation abilities. 
All of our fine-tuning procedure requires only two GPUs, each equipped with 18,176 CUDA cores, 142 streaming multiprocessors, and 48 GB of VRAM, running for approximately seven days—demonstrating greater computational efficiency compared to prior works.

\subsection{Compare with SOTA Methods}
We compare our SPGen with other SOTA works on GSO, Objaverse validation split and Deepfashion3d test split. For GSO and Objaverse validation, we picked the front left view of the object as the condition images. For Deepfashion3d, we follow~\cite{yu2024surf} and pick the front view of sketch or RGB image as conditions.

\subsubsection{Qualitative Results}
As shown in Fig.~\ref{fig:comp_gso}, our generated shapes yield consistent and smooth geometry, this is due to the adoption of SP maps as the coherent representation, in contrast, multiview-based methods such as Wonder3D fail when view inconsistency occurs. Moreover, our method handles shapes with relatively complex topologies, such as porous parts or thin-walled cup structures, while methods relying on SDF for surface extraction fail to accurately generate these structures. It is worth mentioning that our method has good symmetry. When one side of the symmetrical object is occluded, our method accurately restores the unseen geometry.

We also conduct visual comparisons on DeepFashion3D in Fig.~\ref{fig:comp_deepfashion}. For general single image shape creation methods, we adopt front-view RGB image as the condition. For SurfD, we adopt sketch as a condition since they focus on the sketch-to-shape setting. While our method can adopt either RGB or sketch as input in this more challenging setting, no existing SDF-based methods can accurately describe open surfaces. SurfD with the point cloud UDF diffusion model beats those SDF-based methods, and we achieve better results based on UDF, indicating our capability of representing and generating open-surface objects.

Additionally, we compare with image-based representations including nested depth maps from Matryoshka~\cite{richter2018matryoshka} and UV Mapping from~\cite{yan2024object}. We compare the surface quality and geometry details under different image resolutions and as shown in Fig.~\ref{fig:comp_gt_more}, our SP map yields significantly better surface quality with fewer jagged artifacts compared with Matryoshka, and richer geometry details (e.g. the gear area) under the same resolution compared with UV Mappings. 

\begin{figure}[t]
  \centering
  \includegraphics[width=1.0\linewidth]{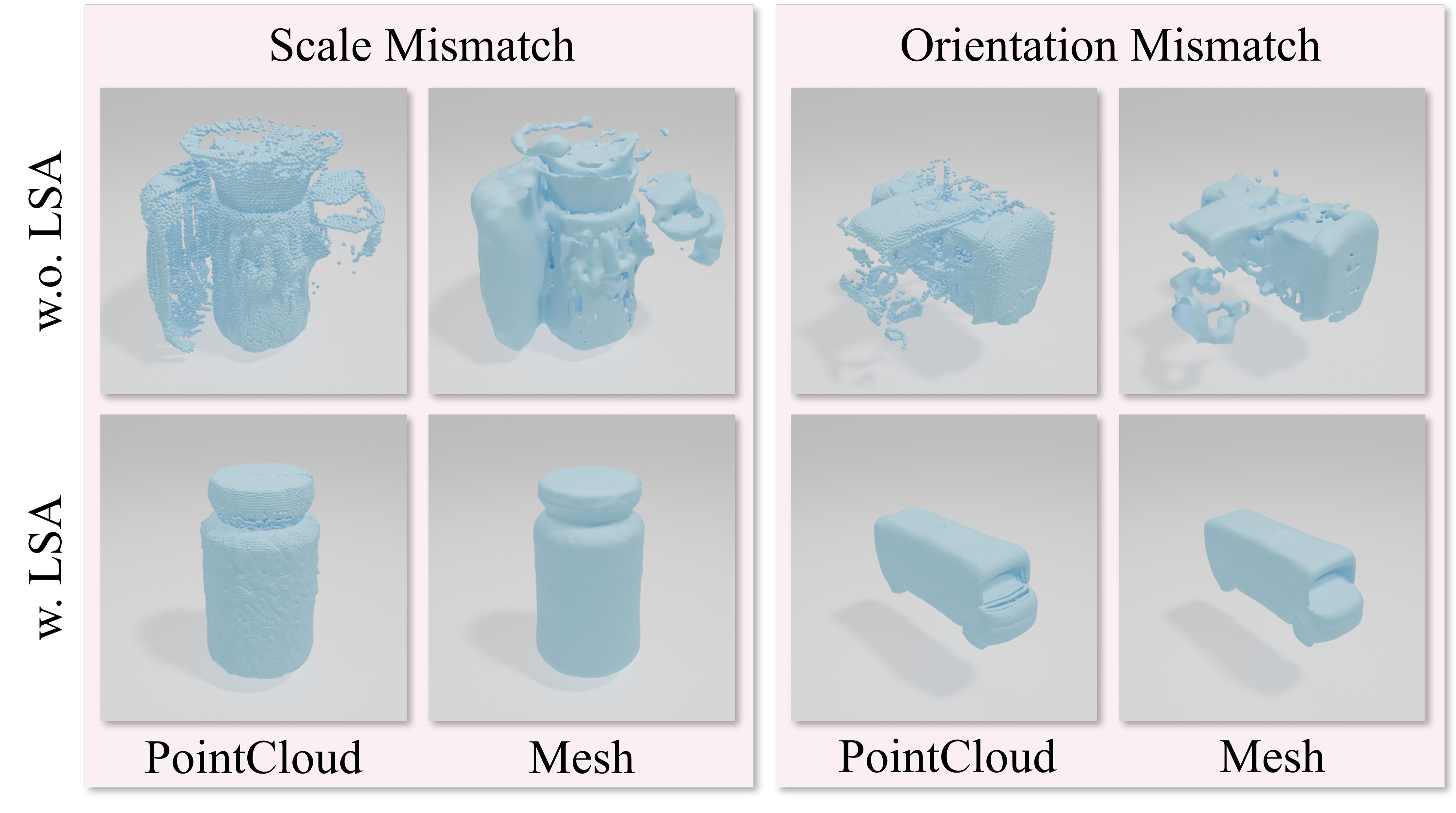}
  \vspace{-6mm}
  \caption{\textbf{Visualization on the effect of Layer-wise Self-Attention (LSA).} Without LSA, the predicted layers show scale or orientation mismatch.} 
  \vspace{-3mm}
  \label{fig:abl_lsa}
\end{figure}

\subsubsection{Quantitative Results}
We use the aforementioned metrics to evaluate the accuracy of generated shapes on both GSO and DeepFashion3d datasets. As shown in Table~\ref{tab:1} and Table~\ref{tab:2}, our method surpasses other SOTA works on all three metrics, and we consume relatively low latency during inference. On GSO dataset, we achieve $+57.5\%$, $+25.4\%$, and $+7.6\%$ relative gain compared with the best performing InstantMesh~\cite{xu2024instantmesh} in CD., Volume IoU and F-Score respectively, and on DeepFashion3D, we achieve $+32.4\%$, $+16.1\%$, and $+8.6\%$ relative gain on the three metrics over SurfD~\cite{yu2024surf}, indicating the effectiveness of our method. We also compare with image-based methods in terms of representation capacity and storage efficiency by reconstructing the ground-truth meshes under different image resolutions. As shown in Table~\ref{tab:2.5}, our SP maps achieve higher reconstruction accuracies and require less storage compared with other two methods under different resolutions, especially at lower resolutions. Since all of these involved 2D image-based representations could be incorporated with the same generative pipeline, our SP Map representation, superior in both reconstruction quality and compactness, provides the generative model with a more accurate and efficient target, allowing higher potential for high-fidelity shape generation.

\subsection{Ablation Study}
\subsubsection{Study on Geometry Regularization}
In Sec.~\ref{sec:losses}, we claimed the importance of applying our proposed geometry regularization to encourage high-quality surface reconstruction and detail preserving during the AutoEncoder training process. Here we conduct ablations to validate the effectiveness. As shown in Fig.~\ref{fig:abs_gr}, we randomly select a small subset with 10k samples from our training split, and train the AutoEncoder on it with or without geometry regularization. Applying geometry regularization greatly speeds up convergence and reduces the training loss from $3.05 \times 10^{-4}$ to $1.47 \times 10^{-4}$, testing loss from $3.59 \times 10^{-4}$ to $1.61 \times 10^{-4}$ by 9k iterations. As shown in the right part, after applying geometry regularization, the high-frequency details are greatly enhanced on both SP maps and reconstructed mesh surfaces.
\vspace{-1mm}

\subsubsection{Study on Layer-wise Self-Attention}
In Sec.~\ref{sec:lsa}, we claim that the Layer-wise Self-Attention (LSA) is proposed to learn the spatial relations of different SP layers. To validate its effect, we removed the LSA layers in the denoise UNet, the results are shown in Table~\ref{tab:3} and Fig.~\ref{fig:abl_lsa}, removing LSA resulting in the mismatch of either layer scale or orientation, and leading to poor mesh quality. 
\vspace{-1mm}

\subsubsection{Study on Finetuning}
As we mentioned in Sec.~\ref{sec:pre}, we finetuned the whole pipeline since the pretrained weights are only generalized on RGB images. In Table.~\ref{tab:3}, we analyze the role of finetuning the pipeline, without finetuning either component of the pipeline, the performance drops significantly, therefore, it is necessary to finetune the whole pipeline on all SP maps.

\begin{figure}[t]
  \centering
  \includegraphics[width=1.0\linewidth]{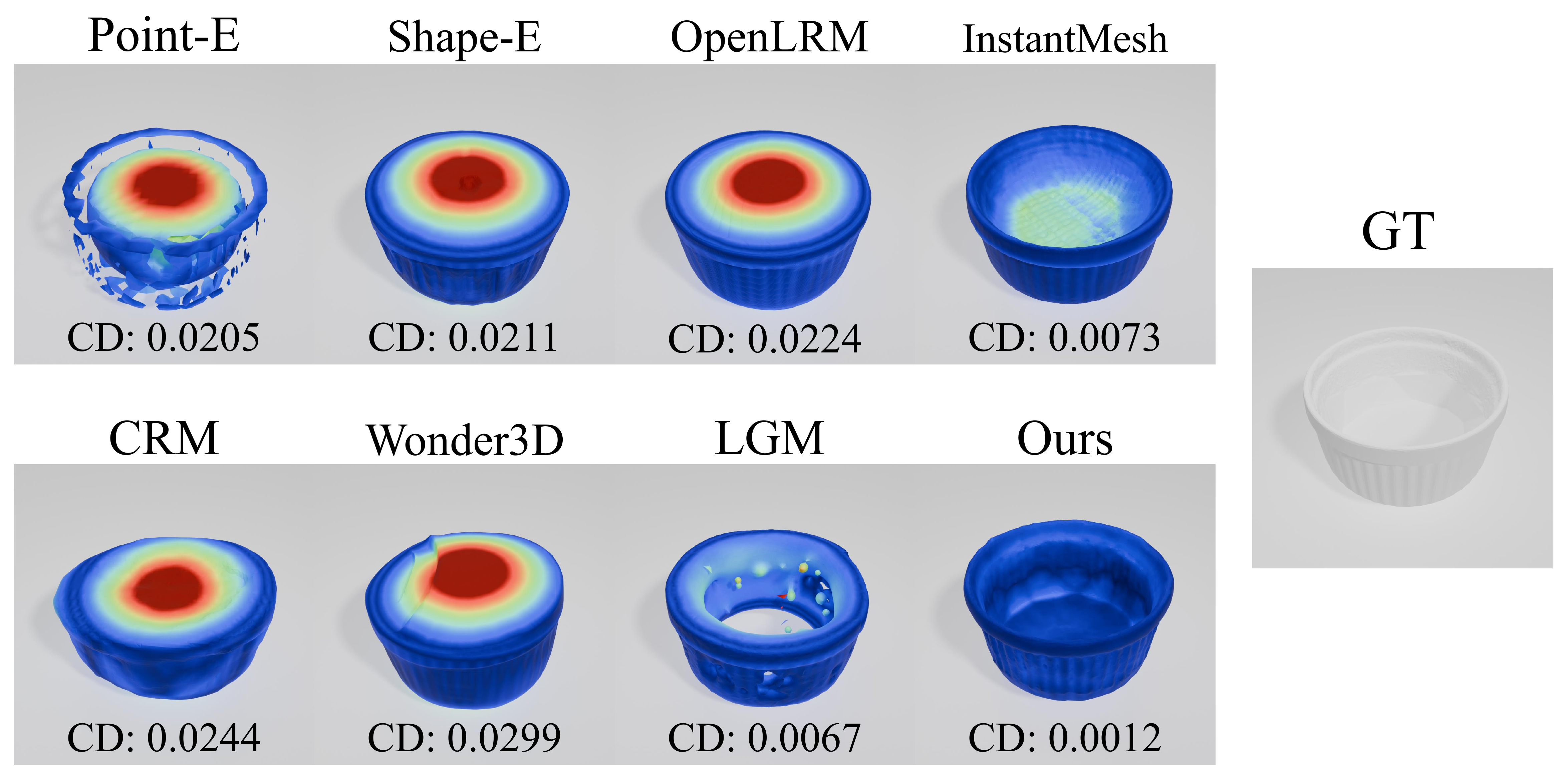}
  \vspace{-6mm}
  \caption{\textbf{Qualitative comparison demonstrating the correlation between Chamfer Distance (CD) and geometric quality.} Lower Chamfer Distance (CD) values correspond directly to higher-fidelity 3D reconstructions with fewer visual artifacts.} 
  \vspace{-3mm}
  \label{fig:abl_cd}
\end{figure}

\subsubsection{Significance of Chamfer Distance Improvements}
We use Chamfer Distance (CD) to quantify the geometric accuracy of the reconstructed meshes. We emphasize that minor numerical reductions in CD correspond to significant and visually perceptible improvements in mesh quality as shown in Fig.~\ref{fig:abl_cd}. A lower CD score directly reflects an enhanced ability to capture correct object structures, maintain surface smoothness, and eliminate artifacts. Thus, the incremental CD gains reported in our experiments represent meaningful progress towards generating high-fidelity 3D geometry.

\section{Conclusions}
\label{sec:conclusions}
We propose SPGen, a novel framework for high-quality 3D shape generation using multi-layer Spherical Projection (SP) as a structural representation. SPGen effectively addresses three key challenges in current 3D creation models: view inconsistency, limited representation capability and low efficiency. By projecting 3D mesh surfaces onto a unit sphere and unfolding them into 2D SP maps, our method ensures geometric consistency through the injective mapping from SP maps to 3D surfaces and flexibly captures complex topologies, including internal layers and open surfaces. SPGen incorporates our proposed geometry regularization and layer-wise self-attention to enhance geometry quality. Extensive experiments demonstrate that SPGen outperforms existing methods in geometric accuracy while maintaining low computational cost and latency, making it a robust and efficient solution for 3D shape generation.

\bibliographystyle{ACM-Reference-Format}
\bibliography{sample-base}


\begin{figure*}[t]
  \centering
   \includegraphics[width=1.0\linewidth]{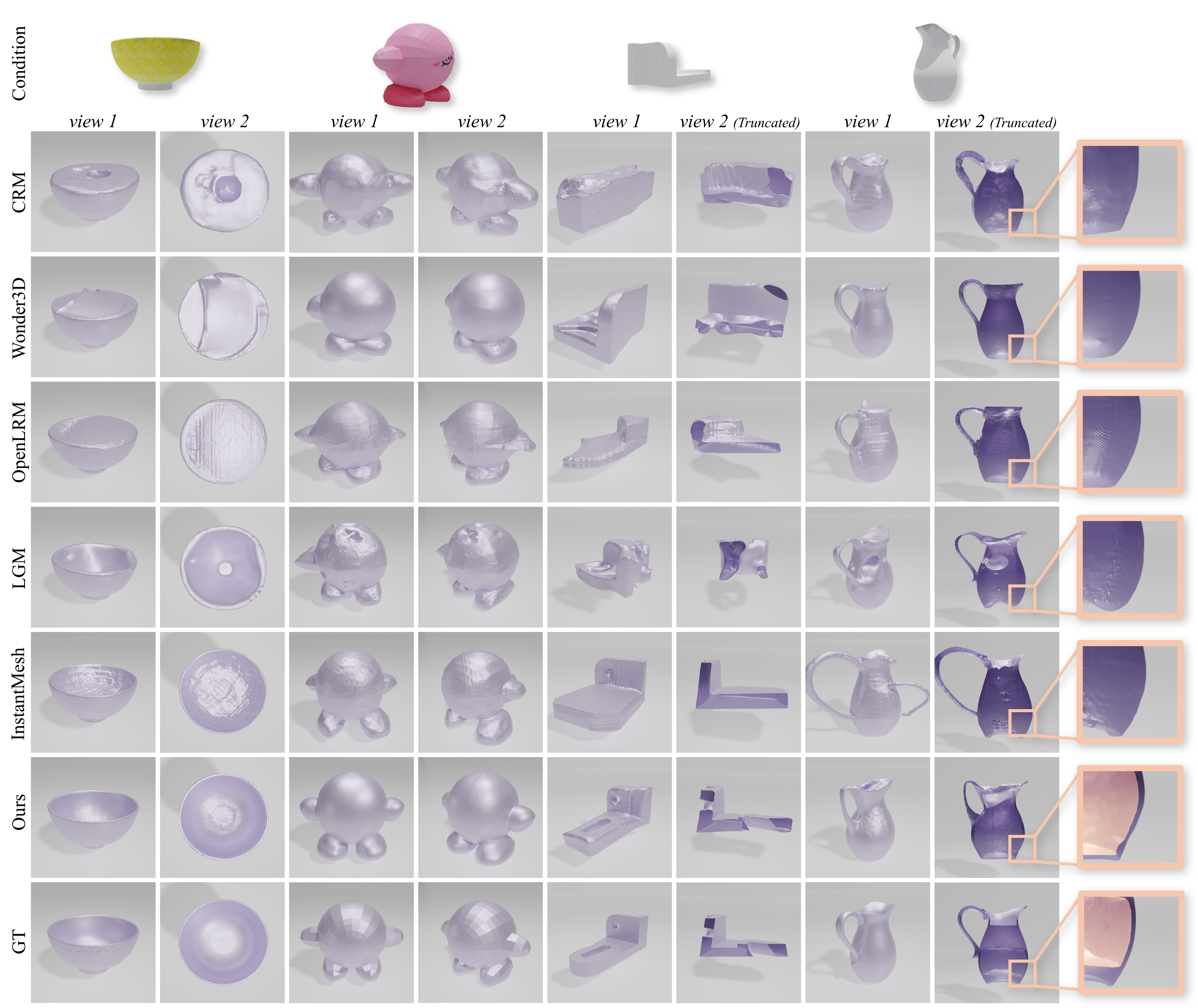}
   \caption{\textbf{Qualitative comparison with SOTA works}. Our SPGen yields highly accurate surface geometry with better topologies.} 
   \vspace{50mm}
   \label{fig:comp_gso}
\end{figure*}

\begin{figure*}[t]
  \centering
   \includegraphics[width=0.85\linewidth]{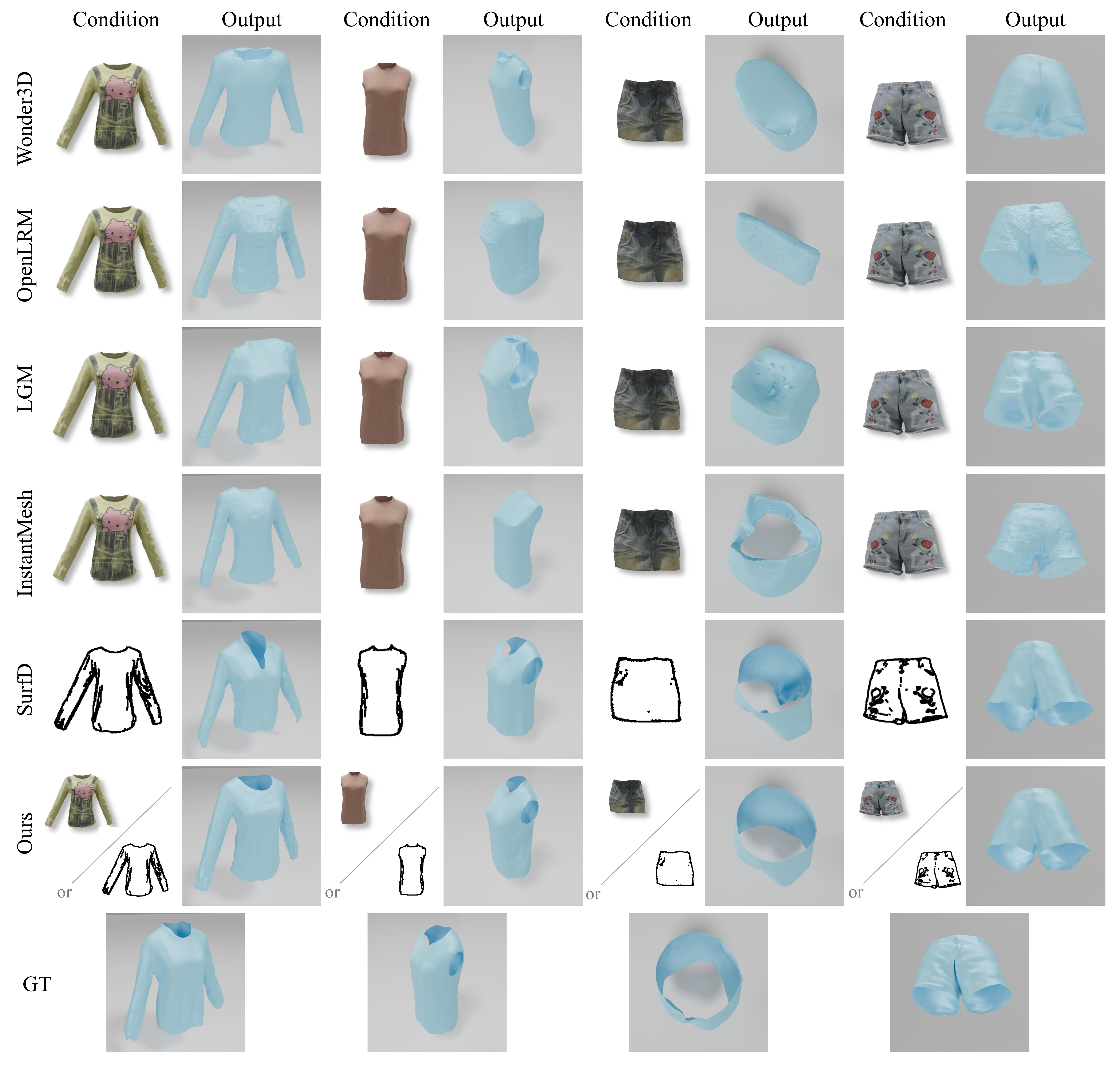}
   \caption{\textbf{Qualitative comparison with SOTA works on DeepFashion3d test split.} Our SPGen can handle either sketch or RGB images individually as the condition. Since the geometric visual effects of the results conditioning by two different conditions are similar, we only show the effect of using RGB as the condition here.} 
   \vspace{10mm}
   \label{fig:comp_deepfashion}
\end{figure*}

\vspace{100mm}

\appendix

\textbf{\huge Supplementary}
\vspace{10mm}

In this supplementary material, we will discuss: i) More details of designs and implementations. ii) More comparisons with image-based geometry representations and other generative pipelines. iii) More ablation studies. iv) Downstream application scenarios. v) Video results showcase.

\section{Detail Design and Implementation}

\begin{figure}[htbp]
  \centering
  \includegraphics[width=1.0\linewidth]{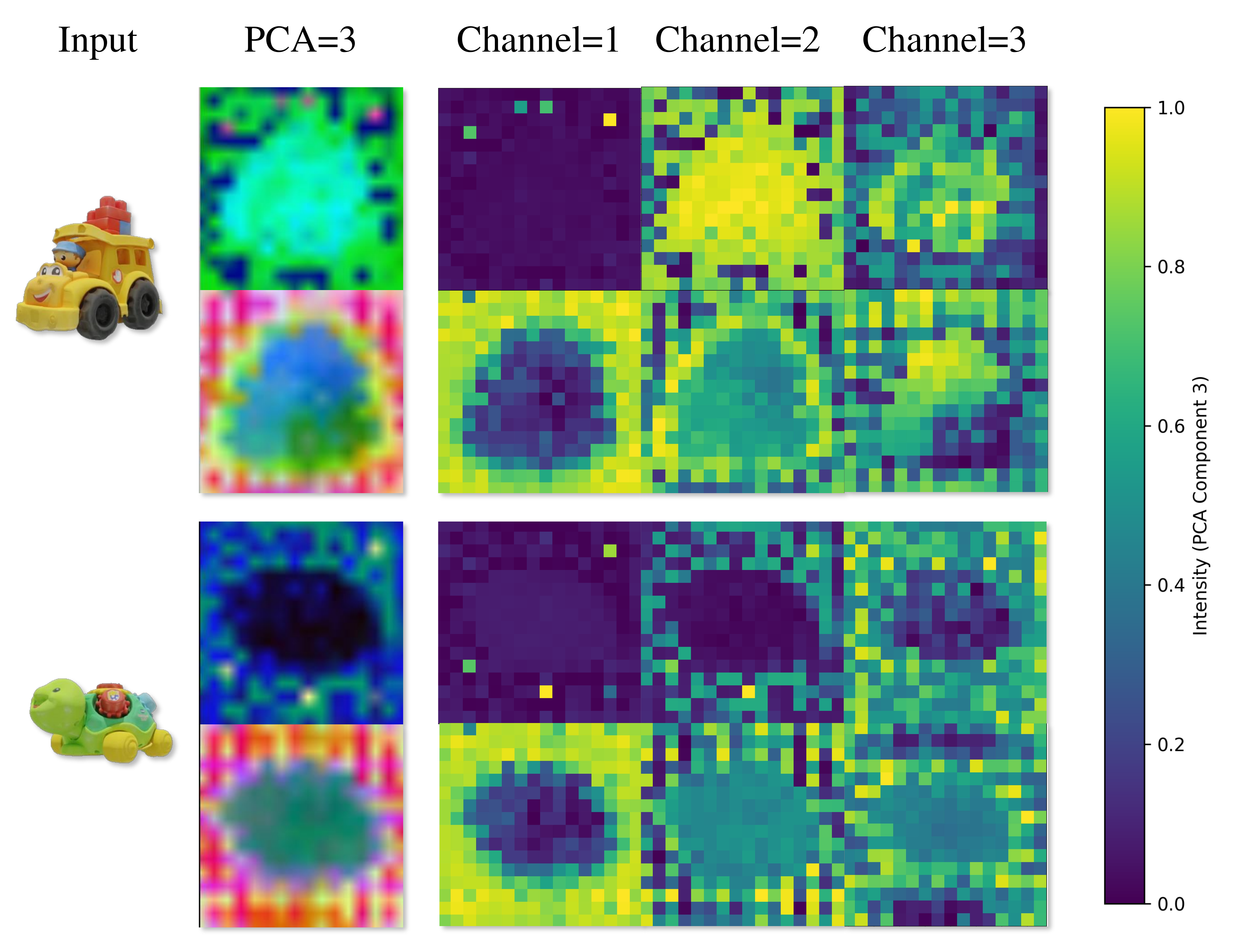}
  \caption{\textbf{Visualization of the visual embedding distributions.} We visualize the final layer token maps of CLIP and DINOv2 via PCA and show the first three main components. The primary component of CLIP feature contains almost no spatial information, and for other major components, the quality of the feature is also lower than DINOv2.
  } 
  \label{fig:comp_score}
\end{figure}

\subsection{Single Image Conditioned Denoising}
\label{sec:cond}
We further claim the usage of single image conditioning and the corresponding visual encoder in this section. To control the denoising process, conditions are usually applied by cross-attention~\cite{vaswani2017attention}, to the UNet layers. For image conditions, usually a pretrained vision encoder is applied to embed the condition image into high dimension feature space, and then perform cross-attention with hidden states of UNet. These vision encoders are trained by contrastive learning~\cite{he2020momentum,radford2021learning} or self-distillation~\cite{caron2021emerging,oquab2023dinov2} to gain the ability to capture image semantics and structures. Current shape generation baselines usually adopt either or both kinds of vision encoders to guide denoising.

We investigate the quality of the visual embeddings produced by the most commonly used contrastive learning CLIP model and self-distillation DINOv2 model. These models use the same ViT~\cite{dosovitskiy2020image} backbone so we take the output token map from the final layer and use principal component analysis (PCA) to reduce the channel dimension to 3. As shown in Fig.~\ref{fig:comp_score}, the primary component from CLIP contains almost no spatial information, since the contrastive learning process focuses more on matching the global semantics of image-text pairs. Differently, DINOv2 adopts various data augmentations to enhance the capturing of image-level details, therefore gaining more refined token maps. For our goal, we target generating 3D shapes that are more similar to the condition image, which requires more image-level details to perform cross-attention and fertilize the 3D shape generation process. Thus, we take DINOv2 as our vision-encoder for diffusion pipeline.

\begin{figure}[t]
  \centering
  \includegraphics[width=1.0\linewidth]{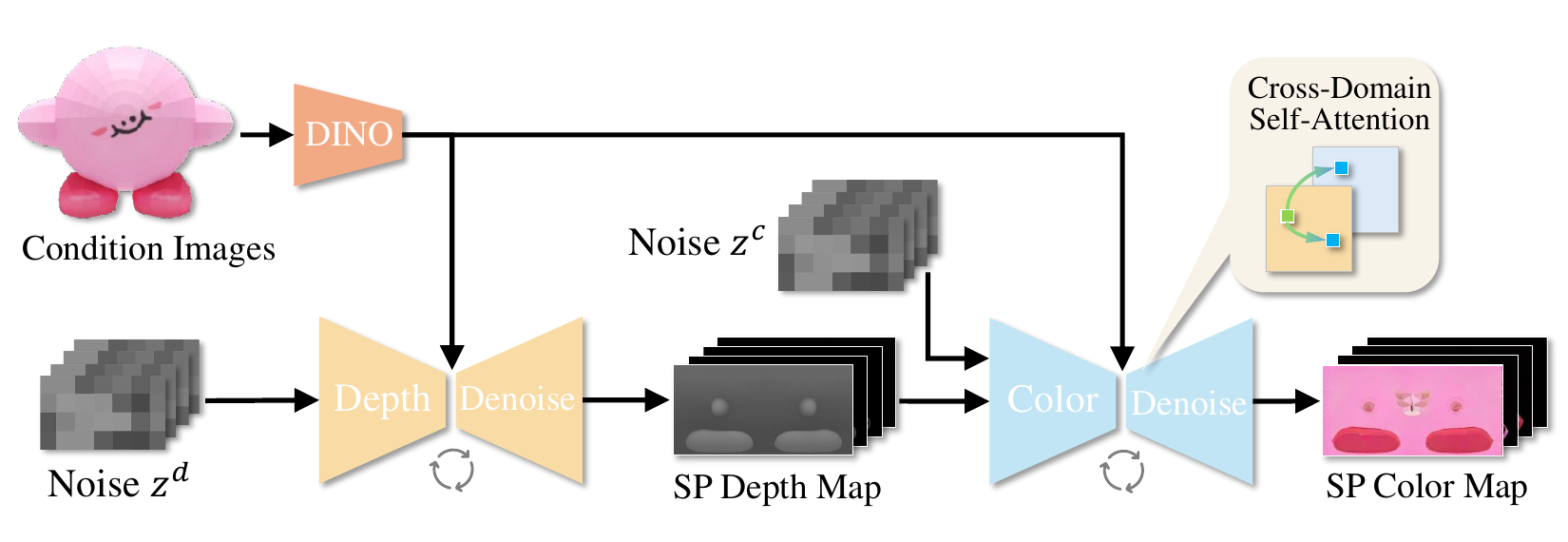}
  \caption{\textbf{Illustration of the depth and color SP maps generation pipeline.} We adopt a cascade pipeline to generate SP depth maps with single-view image condition first, and then use both the generated SP depth map and the single-view image to condition the SP color map generation. The cross-domain self-attention is applied to guarantee the color and depth are matched on the generated maps. Note that we also adopt the latent diffusion pattern for both stages, and the autoencoder is omitted. 
  } 
  \label{fig:sub_arch}
\end{figure}

\subsection{Spherical Projected Texture Map}
Our way of generating Spherical Projection maps to record depth as the geometry can also be extended to different surface attributes, e.g. texture, normal vector, curvature, etc. We show the capability of SPGen on texture generation by recording the vertex colors of each intersection point on an SP color map to represent the surface texture (corresponding with the SP depth map representing the geometry). The map preparation process is the same as we explained in Sec. 3.1. To generate the SP color map, we implement an extra diffusion pipeline conditioned by both the single-view image and SP depth map. As shown in Fig.~\ref{fig:sub_arch}, the whole pipeline is based on a cascade structure, in the depth denoise stage, we use the same diffusion model as in Fig. 2, after the SP depth maps are generated, in the color denoise stage, we use both the SP depth latents and single-view image embeddings to jointly condition another identical denoising UNet.

As discussed in~\cite{long2024wonder3d,wu2024direct,meng2025zero}, multi-attribute images like RGB, depth, normal can be generated in a shared denoising UNet by domain switcher~\cite{long2024wonder3d,meng2025zero} or extra domain-specific branches~\cite{wu2024direct}, and bunches of works have proven the benefits of learning multiple domains together leading to the joint promotions~\cite{radford2021learning,long2024wonder3d,wu2024direct,vandenhende2021multi,vandenhende2020mti,chen2024diffusion,zhang2025bridgenet,phillips2021deep,zhang2024multi}. However, these methods do not apply to SP maps, as the RGB and depth domains on SP maps are significantly distinct. Though domain gaps also remain among different attributes in perspective projected images, they still share the basic image-level structures, i.e., contours, shapes, etc. In contrast, different attributes on SP maps can result in completely distinct map distributions, e.g., a sphere with complex texture or a complex shape with pure color. To avoid negative transfer of knowledge among these domains, we adopt a decoupled pipeline that denoising the shape and texture separately based on their corresponding SP maps. Specifically, following~\cite{shi2023mvdream}, after the SP depth latents $z^d$ are generated, we add noise to them in the set steps $t$ to simulate the noisy distribution $z^d_{t}$, and then feed the noisy depth latent into the color denoising UNet to record the intermediate hidden status. Afterwards, we sample pure noise from a uniform distribution and also feed it into the same UNet, with conducting Cross-Domain Self-Attention with the pre-saved depth hidden status. Thus, the learning of the texture map could be bundled to the corresponding shape. We also apply the embeddings from the single-image condition to provide the visible texture information to the network by cross-attention. Finally, after the denoising and generating $z^c$, we use a specifically finetuned VAE decoder to restore the SP color maps.

\begin{figure}[t]
  \centering
  \includegraphics[width=1.0\linewidth]{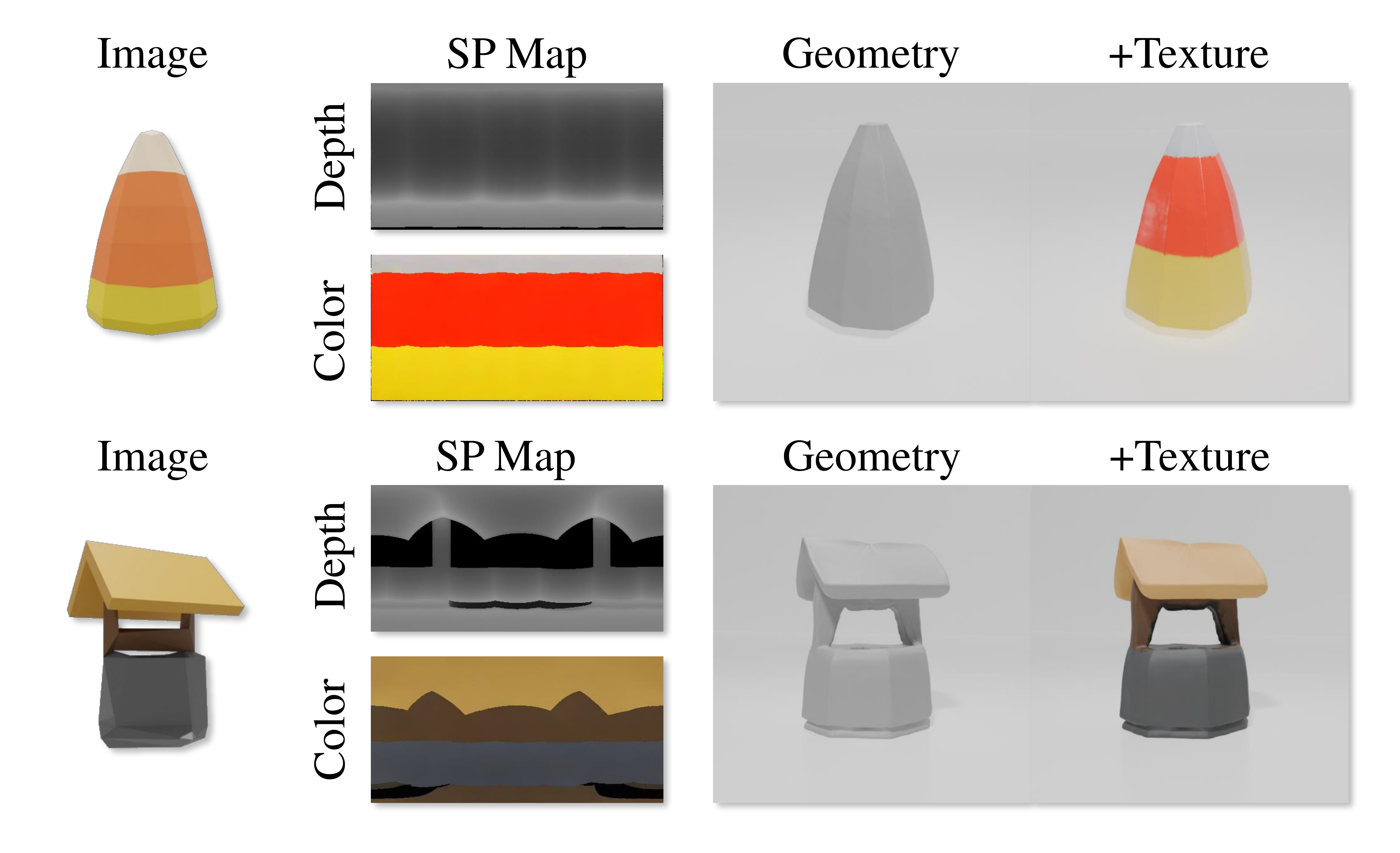}
  \caption{\textbf{Visualization of the textured shape generation.}} 
  \label{fig:text_gen}
\end{figure}

Due to the resource limitation, we conduct a small-scale experiment on a split of 2k Objaverse training data, as shown in Fig.~\ref{fig:text_gen}, the results shows the feasibility of generating textured mesh via Spherical Projection maps.

\begin{table}[t]
\begin{center}
\setlength{\tabcolsep}{7.3mm}{\scalebox{0.75}{
\begin{tabular}{lccc}
 \toprule
{Method}& {CD.}↓ &{Vol. IoU}↑ & {F-Sco. (\%)↑}\\
\midrule
Wonder3D&	0.0246	&	0.3618	&	74.86 \\
CRM	&	0.0172	&	0.3945	&	79.28 \\
OpenLRM	&	0.0136	&	0.4512	&	82.42 \\
LGM	&	0.0203	&	0.3756	&	77.59 \\
InstantMesh	&0.0157		&0.4108	&	82.62 \\ \midrule
Ours	&\textbf{0.0061}		&\textbf{0.5527}		&\textbf{94.74} \\
\bottomrule
\end{tabular}
}}
\end{center}
 \caption{\textbf{Quantitative comparison on Objaverse validation split.}}
\label{tab:s1}
\end{table}

\begin{table*}[t]
\centering
{
  \setlength{\tabcolsep}{5.3mm}   
  \scalebox{0.75}{              
    \begin{tabular}{l|ccccc|ccc}
      \toprule
      Method & GPU & Training Time & Iterations & Data Amount & Latency
             & CD.\,$\downarrow$ & Vol.\ IoU\,$\uparrow$ & F-Score (\%)$\uparrow$\\
      \midrule
      CLAY        & 256 \textit{A800} (10 TB)    & $\sim$2 \textit{weeks} & –    & 527k & $\sim$15s & 0.0046 & 0.6355 & 96.95 \\
      Trellis           & 64  \textit{A100} (2.5 TB)   & –                     & 400k & 500k & $\sim$40s & 0.0030 & 0.6495 & 98.35 \\
      Hunyuan3D-2       & –                    & –                     & –    & –    & $\sim$15s & 0.0028 & 0.7440 & 98.43 \\
      TripoSG           & 160 \textit{A100} (6.25 TB)   & $\sim$3 \textit{weeks}& 700k & 2m   & $\sim$50s & 0.0030 & 0.7381 & 99.08 \\
      TripoSF           & 64 \textit{A100} (2.5 TB)    & - & - & 400k   & - & - & - & - \\
      \midrule
      Ours              & 2  GPUs (0.09 TB) & $\sim$1 \textit{week} & 80k  & 160k & 6–10s     & 0.0034 & 0.6208 & 98.28 \\
      \bottomrule
    \end{tabular}
  }
}
\caption{\textbf{Quantitative comparison with large foundation 3D generative pipelines.} }
\label{tab:s2}
\end{table*}

\subsection{Implementation Details}
\subsubsection{Dataset Preparation.} We refer to the criteria in~\cite{chen2024meshanything,long2024wonder3d} to filter the Objaverse dataset by removing low-quality or scene-level meshes and acquire around 160k objects as our whole training split. We also picked 1993 objects out of the training indices as our validation split on Objaverse. Before training, we follow~\cite{long2024wonder3d} and render 13 views per object with resolution $512 \times 512$ including orthographic and other oblique perspectives by Blender~\cite{blender2018blender}. We normalize the scales of object meshes to the range of $\left[-0.5, 0.5\right]$ without normalizing the orientations to maintain diversity, and translate objects to the origin. We also fix the sphere-center at the origin which is used to scan SP maps. The azimuth range for SP map is $\theta \in [-\pi / 2, 3\pi / 2)$, and the polar range is $\varphi \in [0, \pi)$.

\subsubsection{Model Details.} We use the VAE and UNet from SDXL~\cite{podell2023sdxl}, and load the pretrained weight of both to conduct finetuning. For VAE reconstruction, we apply L1 loss and our geometry regularization on the reconstructed maps, and a KL-divergence regularization with the weight of $10^{-8}$ to ensure the latent space won't shift away from the uniform distribution during the finetuning. For the denoising Unet, we apply LSA layers and adopt L2 loss in the latent space. After the SP maps are generated, we unproject the points on map into the 3D space, and use a normal estimator to predict normal vector for each point. The normal estimator is a ConvONet~\cite{peng2020convolutional}, we scale it up to $192$ hidden dimensions, and the 3D-UNet encoder contains 5-levels with 128 feature dimensions. And the model is also trained on all our training split by sampling $25600$ oriented points on each mesh object. 

\begin{figure}[t]
  \centering
  \includegraphics[width=1.0\linewidth]{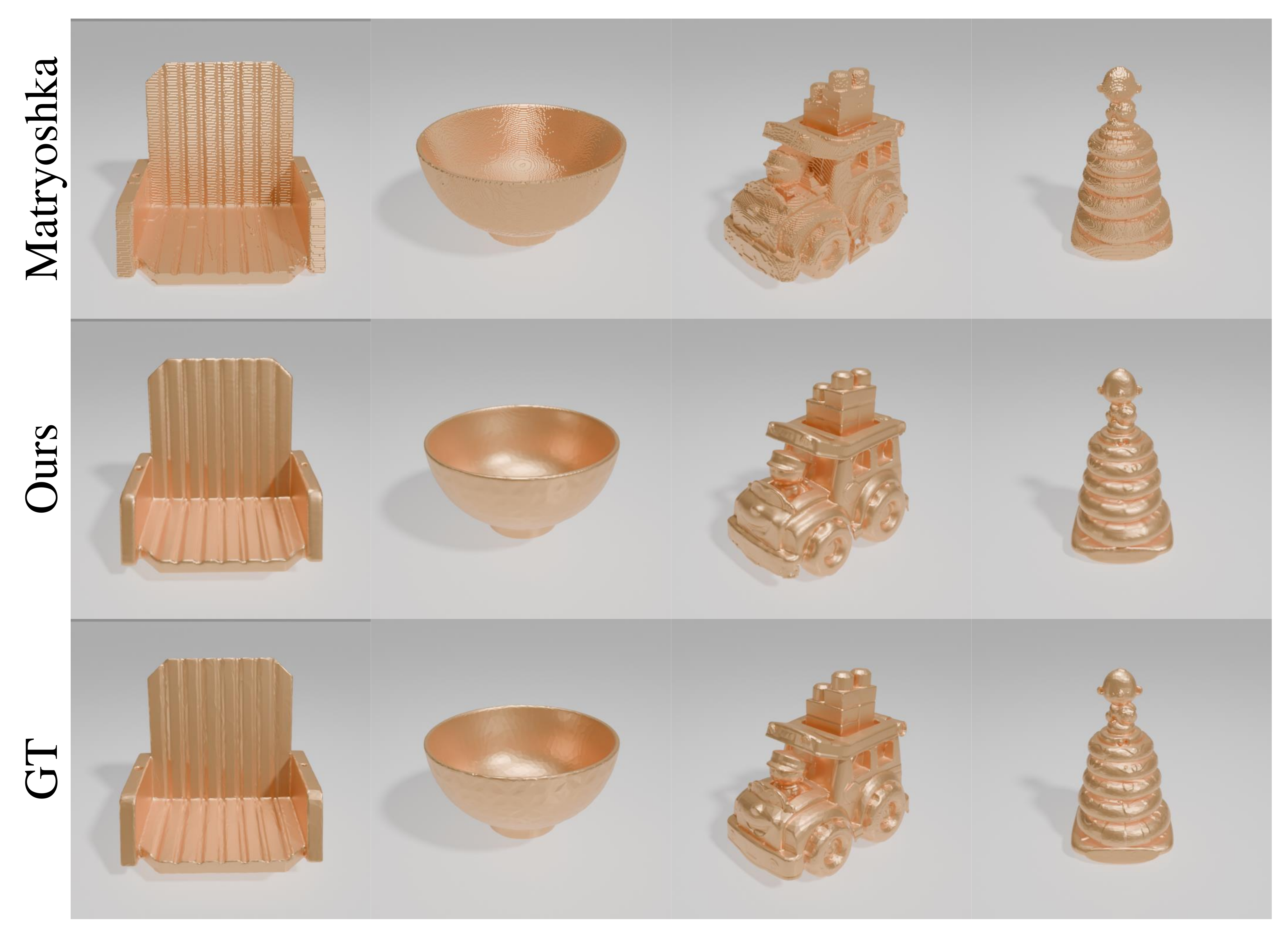}
  \caption{\textbf{Visualization reconstruction quality of Matryoshka~\cite{richter2018matryoshka} and our SP maps.} We reconstruct the ground-truth directly with both methods, the comparison shows that our SP maps maintains better surface details compared with matryoshka.
  } 
  \label{fig:comp_gt}
\end{figure}

\subsection{Limitations}
\subsubsection{Faces Parallel to the Ray Direction.}
A theoretical ambiguity may arise when surface faces are exactly parallel to the sphere radius (i.e., orthogonal to the spherical surface), as in the flat boundary of a hemisphere. In such cases, rays shot from the sphere center may intersect the face tangentially or fail to yield a unique depth, leading to potential instability. Our implementation is based on the Möller–Trumbore algorithm~\cite{moller1997fast} and explicitly detects near-parallel ray-face configurations by recording the angle between the ray and face normal. When the angle falls below a small threshold, the intersection is discarded to avoid numeral unstable cases. Empirically, such events are usually rare—fewer across the dataset, and in pathological cases with large flat regions aligned with the radial direction, we have to apply a small random perturbation to the ray origin or surface to ensure numerical stability.

\subsubsection{Distortions at the Polar Areas.}
Equirectangular projection of the sphere introduces non-uniform sampling, especially near the polar regions, which may cause distortion, which could possibly lead to inaccurate geometry details in the polar areas. However, because SP maps have no severe degeneration or flipped mapping, these distortions remain acceptable and errors on 3D surfaces are controllable. We conduct analysis in the ablation study.

\begin{figure*}[t]
  \centering
   \includegraphics[width=1.0\linewidth]{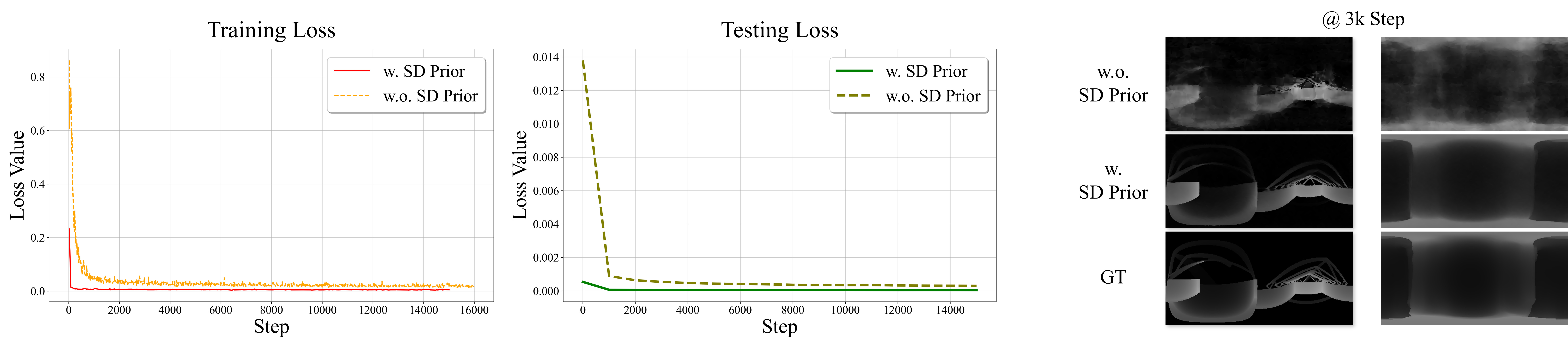}
   \caption{\textbf{Analysis on SD priors.} On the left, we compare the loss curve on training and testing splits, and on the right we visualize comparison of SP depth maps at step 3k with and without SD priors.} 
   \label{fig:abs_sdp}
\end{figure*}

\begin{figure}[t]
  \centering
  \includegraphics[width=1.0\linewidth]{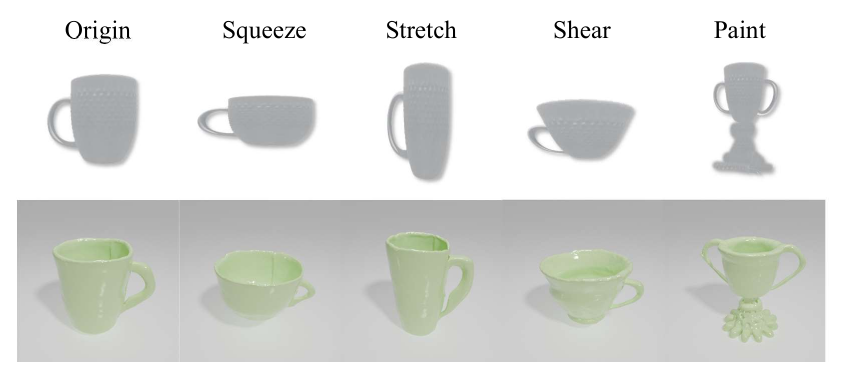}
  \caption{\textbf{Visualization of the shape editing via image control.} We try to edit the original image and SPGen is able to generate corresponding shapes accurately.
  } 
  \label{fig:comp_edit}
\end{figure}

\begin{figure}[t]
  \centering
  \includegraphics[width=1.0\linewidth]{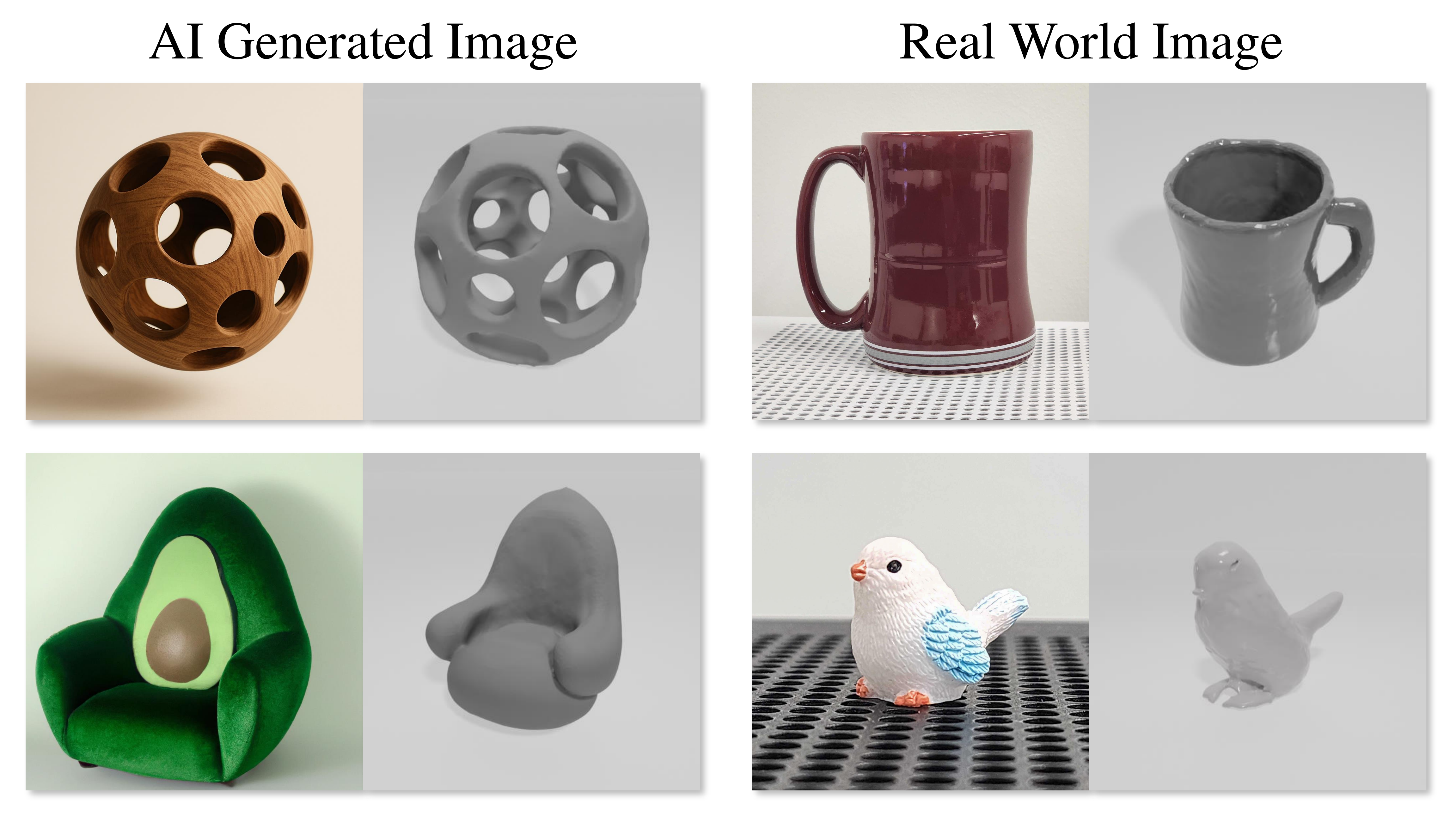}
  \caption{\textbf{Real-world Evaluation.}
  } 
  \label{fig:vis_wild}
\end{figure}

\section{Comparisons and Discussions}

\subsection{Comparisons on Objaverse Validation Split}
As we mentioned in Sec. 4.1.1, we set a small validation split on Objaverse with 1993 objects to indicating the status of model convergence. Note that since different works are adopting different data filtering strategy on Objaverse, so our validation data could be the training data of other works. We conduct quantitative comparisons with our baselines on this split. We randomly choose 20 shapes consisting of common objects used in daily life, and the results are shown in Table~\ref{tab:s1}. Our SPGen also achieves consistent gain on all three metrics compared with other works.

\subsection{Comparisons with Image-based Geometry Representations}
\subsubsection{Comparison with Zero-1-to-G}
Zero-1-to-G~\cite{meng2025zero} is an extended multi-view diffusion method from~\cite{long2024wonder3d} incorporating multiple Gaussian attributes, whereas SPGen differently uses consistent spherical projections and directly encodes surface geometry. Zero-1-to-G cannot guarantee the strict view-consistency and relies on an extra differentiable-rendering stage for surface extraction.

\subsubsection{Comparison with Geometry Image and UV Mapping}
Geometry Image~\cite{gu2002geometry,elizarov2024geometry} is a representation that unfolds a mesh surface onto a single regular 2D grid so each pixel stores the surface’s $(x,y,z)$ coordinates, while UV Mapping~\cite{yan2024object} assigns 2D $(u,v)$ coordinates to a parametrized mesh surface to accurately wrapped over it. Both representations unfold the 3D surface onto a structural 2D domain which makes it possible to leverage the strong priors from pretrained image generative models. However, these representations are different from SP maps in: 1) These methods have non-unique geometry mapping of the same object, which hinders the construction of consistent large-scale datasets and scalable model training; 2) Both of them require extensive cutting to unfold mesh surface, especially on objects with genus $>0$, which potentially leads to more burdens for model to learn the boundary relations, and possibly leads to higher border errors with unstitched patches (refer to Fig.5 in~\cite{yan2024object} and Fig.7 in~\cite{elizarov2024geometry}). In contrast, SP maps leverage fixed mappings and cuts, which ensure the building a standardized and scalable generative training pipeline.

\subsubsection{Comparison with Matryoshka Network}
Matryoshka Net-\\
work~\cite{richter2018matryoshka} encodes a shape by predicting six axis-aligned stacks of nested depth images (one for each $\pm X,\pm Y,\pm Z$ direction) that are fused into a fixed-resolution voxel grid before polygonization. This design can also effectively record 3D shapes in structural 2D images and takes advantage of image processing networks such as 2D CNNs. However, compared with SP maps, it is still limited in (i) the voxel–resolution bottleneck: though the nested depth images are $N^{2}$ complexity, the surface detail is capped by the $N^{3}$ volume, whereas SPGen reconstructs geometry directly from high-resolution spherical-projection maps whose memory grows only quadratically, enabling lower budgets in reconstructing the geometries; and (ii) the cross-view or cross-layer consistency: since no explicit cross-attentions among views or layers to ensure the consistency, though the final depth fusion step will store a unique shape, the depth stacks may still contain inconsistency (e.g., entry/exit order inversion along a ray, misaligned strip boundaries), which leads to possible holes, thinned walls, or jagged artifacts despite on the restored shape. In contrast, SP map is a naturally view consistent injective function plus layer-wise self-attention to eliminate potential conflicts.

We also conduct ground-truth surface reconstruction experiments on both Matryoshka and SP maps, we still use $256 \time 512$ as the resolution for SP maps, and $4$ layers in depth. To align with our setting, we use $256^3$ of spatial resolution for Matryoshka, and also set maximum $4$ layers. Note that $256^3$ voxel is relatively expensive during surface extraction since the original setting of Matryoshka in single-view reconstruction is $32^3$. As shown in Fig.~\ref{fig:comp_gt}, our reconstruction results achieve higher quality with more details, while Matryoshka results are suffering from jagged artifacts and coarse surfaces. 

\subsubsection{Comparison with GenRe}
 GenRe~\cite{zhang2018learning} adopts a cascaded, multi-stage pipeline that first predicts a depth map from an RGB image, projects it to a partial spherical map, then completes this map using a feed-forward 2D inpainting network, and finally projects the result to a voxel grid for refinement. This pipeline has key limitations that our end-to-end framework effectively addresses i) Representation capability: GenRe uses a single-layer spherical map that lacks internal structure representation and relies on post voxel-based refinement, leading to resolution bottleneck. In contrast, our multi-layer SP maps efficiently capture complex topologies directly in the 2D domain. ii) External dependencies: GenRe relies on predefined camera parameters and a separate depth estimato, which introduces error accumulation. Our framework is fully self-contained and free from such dependencies.

\subsection{Comparisons with More 3D Generative Pipelines}
Recently, there are bunches of scalable 3D generative models trained on large-scale datasets yielding strong generalization ability and robustness on generating high-quality 3D meshes. These works adopt 3DShape2VecSet~\cite{zhang20233dshape2vecset} (\cite{zhang2024clay,li2025triposg,zhao2025hunyuan3d}) or Sparse Voxel (~\cite{xiang2024structured,he2025sparseflex}) as the geometry representation, and use signed distance functions or occupancy field as implicit surfaces. These works usually adapt scalable DiT~\cite{peebles2023scalable} and train on $\sim 500$k or more data. As shown in Table~\ref{tab:s2}, we compare to these works on randomly selected GSO and Objaverse-validation data. Since CLAY is a closed-source work, we use their API Rodin Gen-1. Our SPGen only consumes less than $5\%$ of the training resources to achieve relatively competitive performance with faster inference speed, indicating the compactness and effectiveness. 


\section{Ablation Studies}
\subsection{Study on Pretrained SD Priors}
Recent studies have demonstrated that SD models are highly adaptable and can improve performance on various 2D representations beyond RGB, including panoramic images~\cite{wang2023360}, depth and Gaussian feature maps~\cite{wu2024direct,yu2025gaia,meng2025zero}, normal maps~\cite{long2024wonder3d}, etc. These works show that SD priors are beneficial in bridging domain gaps and performing generalization. Similar to~\cite{wang2023360,wu2024direct}, our SP map is unfolded surface depth image with panoramic-style distortion, retaining image-level local structures and spatial patterns, which can be naturally benefited from SD priors. We also conduct ablation studies on the training convergence with and without SD priors. As shown in Fig.~\ref{fig:abs_sdp}, the model with SD priors achieves significantly faster convergence and lower loss ($4.37 \times 10^{-5}$ \textit{v.s.} $3.08 \times 10^{-4}$ at 15k step on test data), further validating the effectiveness of SD priors post-adaptation. Additionally, we also show the generated SP depth map at training step 3k, with SD Prior, the result has significantly better quality with less noise.  

\subsection{Study on Border Consistency}
We follow~\cite{wang2023360} and apply circular padding in the azimuth direction to ensure the SP maps align well at the border and the rotational invariance. The circular padding encourages the generative model to learn consistent predictions across the entire SP map. We calculate the absolute-relative-error between the azimuth borders, which is only $0.23\%$, which proves the consistent learning effect on borders.

\subsection{Study on the Number of SP Map Layers}
To ensure representation comprehensiveness, we evaluated reconstructed IoU on 160K Objavese objects using different numbers of SP map layers. The results are:

$\left\{1:92.0\%, 2:98.7\%, 3:99.8\%, 4:99.9\%, 5:99.9\% \right\}$, 

\noindent indicating that using less than $4$ layers yields incomplete reconstructions, while additional layers provide negligible improvement. Thus, we select $4$ layers as they effectively model complex objects while balancing representation completeness and computational efficiency.

\subsection{Study on the Distortions.}
We empirically measure the absolute-relative-error across surface regions and observe only marginal variation: $0.20\%$ in polar areas, $0.25\%$ near the equator, and an average of $0.22\%$. This validates that the model handles projection distortion effectively, and that geometric reconstruction from SP maps remains robust even in the presence of polar distortions.

\subsection{Real-World Evaluation}
We evaluated our SPGen with real-world data to verify its generalization ability and robustness. As shown in Fig.~\ref{fig:vis_wild}, we use AI-generated images and daily photos as conditions, our method achieves good geometric quality and restores the conditional image well.

\section{Downstream Applications}
Our generation pipeline can be used for a lot of downstream tasks such as editing, rendering, animation, etc. In this section, we test the editing ability of shapes via the single image control. As shown in Fig.~\ref{fig:comp_edit}, we perform several editing ways including squeeze, stretch, shear, and direct paint on it, our SPGen shows stable and controllable transformation according to the change of input view, indicating the power of robust generation and generalization.

\section{Video Results}
More detailed evaluations and comparisons can be found in the attached video.

\end{document}